\documentclass[journal]{IEEEtran}

\usepackage{fontspec}
\usepackage{amsmath,amssymb}
\usepackage{booktabs}
\usepackage{array}
\usepackage{tabularx}
\usepackage{graphicx}
\usepackage{newunicodechar}
\usepackage{enumitem}
\usepackage{stfloats}
\usepackage[hidelinks]{hyperref}
\usepackage{microtype}

\newunicodechar{²}{\textsuperscript{2}}
\newunicodechar{³}{\textsuperscript{3}}
\newunicodechar{×}{\ensuremath{\times}}
\newunicodechar{·}{\ensuremath{\cdot}}
\newunicodechar{−}{\ensuremath{-}}
\newunicodechar{±}{\ensuremath{\pm}}
\newunicodechar{≈}{\ensuremath{\approx}}
\newunicodechar{≥}{\ensuremath{\geq}}
\newunicodechar{≤}{\ensuremath{\leq}}
\newunicodechar{∈}{\ensuremath{\in}}
\newunicodechar{√}{\ensuremath{\surd}}
\newunicodechar{∼}{\ensuremath{\sim}}
\newunicodechar{∇}{\ensuremath{\nabla}}
\newunicodechar{→}{\ensuremath{\rightarrow}\allowbreak}
\newunicodechar{↑}{\ensuremath{\uparrow}}
\newunicodechar{′}{\ensuremath{'}}
\newunicodechar{∗}{\ensuremath{\ast}}
\newunicodechar{α}{\ensuremath{\alpha}}
\newunicodechar{γ}{\ensuremath{\gamma}}
\newunicodechar{μ}{\ensuremath{\mu}}
\newunicodechar{σ}{\ensuremath{\sigma}}
\newunicodechar{Δ}{\ensuremath{\Delta}}
\newunicodechar{‖}{\ensuremath{\|}}
\newunicodechar{†}{\ensuremath{\dagger}}
\newunicodechar{‡}{\ensuremath{\ddagger}}
\newunicodechar{¶}{\P}
\newunicodechar{§}{\S}
\newunicodechar{…}{\ldots}
\newunicodechar{①}{(1)}
\newunicodechar{②}{(2)}
\newunicodechar{③}{(3)}
\newunicodechar{④}{(4)}

\setlist[itemize]{leftmargin=1.1em,topsep=2pt,itemsep=1pt,parsep=0pt}
\setlist[enumerate]{leftmargin=1.3em,topsep=2pt,itemsep=1pt,parsep=0pt}
\providecommand{\tightlist}{\setlength{\itemsep}{0pt}\setlength{\parskip}{0pt}}

\begin{document}

\title{YOLO26-RD: An End-to-End Road Damage Detection Network With Learnable Contrast Enhancement and Edge-Guided Downsampling}

\author{Sompote Youwai, Pawarotorn Chaipetch, Hathairat Samaikul,
and Theerayut Yonseng%
\thanks{S. Youwai is with the AI Research Group, Department of Civil
Engineering, King Mongkut's University of Technology Thonburi, Bangkok,
Thailand (e-mail: sompote.you@kmutt.ac.th).}%
\thanks{P. Chaipetch, H. Samaikul and T. Yonseng are with Infraplus Co.,
Ltd., Thailand.}}

\maketitle

\begin{abstract}
Pavement-distress detectors are conventionally specialised for small objects, typically by adding a stride-4 detection head and replacing strided convolution with lossless space-to-depth downsampling. This paper measures that premise against the annotation geometry of the target data and shows that it does not hold for region-level survey imagery: 1.28\% of instances are small at the 640² training resolution while 70.37\% are large, yet a stride-4 level would claim 75.3\% of all anchors, and complete misses rather than localization errors dominate the failure decomposition of a trained baseline. Guided by that audit, the proposed detector YOLO26-RD reallocates the anchor budget, retaining the stride-4 branch as neck features while carrying no detection level there, and adds LearnableContrast, a 494-parameter per-tile correction learned from the detection loss and active at inference, and EdgeSPD, a lossless space-to-depth downsampler gated by a fixed Sobel prior. Fifteen models were trained from scratch under one recipe, five scales each of the proposed architecture and of recipe-matched YOLO26 and YOLOv12 families. YOLO26-RD outperforms both reference families. Averaged over the five scales it returns 0.790 mAP50 and 0.482 mAP50-95 against 0.776 and 0.471 for YOLO26 and 0.755 and 0.468 for YOLOv12; it gives the higher mAP50 than both at every scale from s upward, and at scales m, l and x it leads both on mAP50 and mAP50-95 alike, twelve pairwise comparisons decided without exception. YOLO26-RD-l is the best model of the fifteen on both metrics, at 0.809 mAP50 and 0.497 mAP50-95, improving on the YOLO26 reference by 0.031 and 0.030 and leading in all six per-class entries, and every arm of a module ablation exceeds that baseline. The margin is therefore a property of the architecture rather than of one tuned configuration, with the qualifications that three of the twelve scale-wise margins lie inside the dataset's ±0.015 resolution limit and that the held-out split reproduces the ordering against YOLO26 but not against YOLOv12 at the largest scale. As a TensorRT FP16 engine the released model sustains 98 frames per second on an entry-level accelerator against the 21 a 100 km/h survey requires.
\end{abstract}


\section{Introduction}\label{introduction}

Road-surface distress, comprising alligator (fatigue) cracking, linear cracks and patch repairs, is a primary input to pavement-management systems, which use it to schedule maintenance and allocate budget across a road network. The imagery analysed here is acquired by a camera fixed above the pavement and oriented perpendicular to it, so each frame is a near-nadir view at approximately known scale, recorded continuously as the vehicle travels. That geometry fixes the viewing angle, removes the perspective foreshortening of forward-facing dashcam datasets, and permits a pixel measurement to be converted to a physical extent. It also determines the data volume, since frames accumulate at the rate the vehicle covers ground rather than at the rate they can be examined. The industrial operator of this system, Infraplus Co., Ltd.~(the second author's affiliation), supplied the imagery and annotations and has until now processed the output by human inspection. The company surveys more than 10,000 km of road in Thailand, which yields far more frames than a review team can examine; manual inspection is accurate but does not scale, and the boundary of a distress region is a judgement that varies between reviewers. Automating the detection step is therefore a precondition for operating the system at network scale.

The form that automation takes determines what the detector has to be. Figure 1 gives the concept. Frames are captured continuously, each is processed on board as it arrives, and the detections are reduced to one structured record per frame carrying the class, box, confidence, timestamp and position of every distress found. Only those records leave the vehicle, returned to the control centre as a damage report feeding the pavement-management system, while the imagery stays on board so that annotations can be revised and models retrained later. The reason is volume: a frame averages 661 KB whereas a record describing its contents occupies about 150 bytes, a reduction of roughly three orders of magnitude. Two requirements follow, and both fall on the detector. It must run inside the vehicle on hardware a fleet can afford to install and power, and it must consume frames at the rate the camera produces them. Real-time here is therefore defined by the road rather than by convention, and Section 8 derives it as roughly 21 frames per second at 100 km/h with consecutive frames overlapping by half.

\begin{figure*}[!t]
\centering
\includegraphics[width=0.86\textwidth]{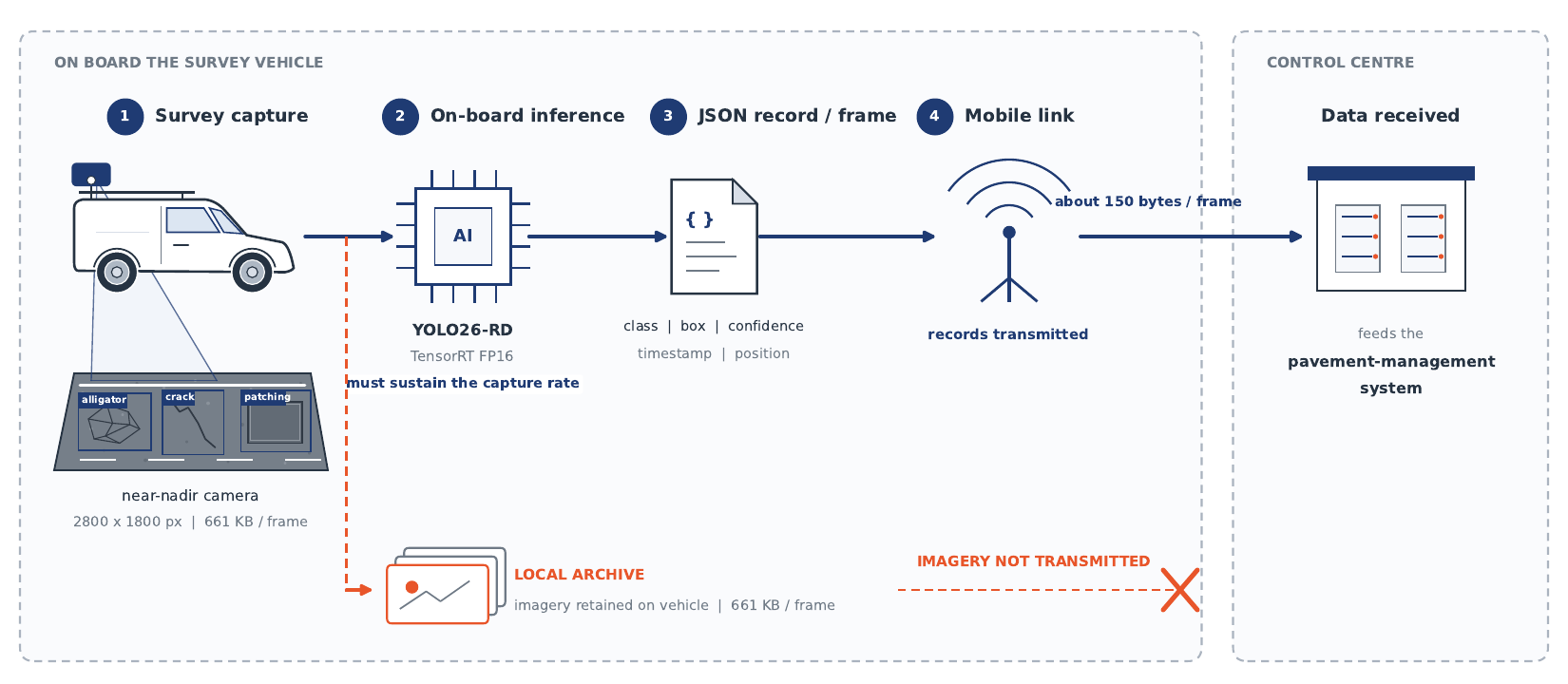}
\caption{Concept of the automated survey. Frames are captured continuously, processed on board as they arrive, and reduced to one structured record each; only those records are transmitted, while the imagery is retained in the vehicle. The detector must therefore sustain the capture rate on hardware carried in the vehicle. Quantities are given in Section 8.}
\label{fig:1}
\end{figure*}
Automation of this kind is dominated by the YOLO family of single-stage detectors, which return a list of boxes each labelled with a class and a confidence, directly comparable to what a human reviewer produces. Four stages of that lineage matter here: anchor-based heads {[}1{]} gave way to anchor-free regression {[}2{]}, multi-scale feature aggregation was restructured by BiFPN {[}3{]} and later by YOLOv9 {[}4{]}, attention became a core backbone component in YOLOv12 {[}5{]}, and non-maximum suppression was eliminated, YOLOv10 {[}6{]} introducing consistent dual assignment and YOLO26 {[}7{]} carrying this into an end-to-end one-to-one head. Every stage was tuned on COCO, a dataset of everyday objects. Road-damage detection inherits both that lineage and its benchmark assumptions: the pavement literature adopts whichever mainstream YOLO is current and modifies it for cracks, following a characteristic recipe whose elements address the same perceived problem, namely that a crack is a small and faint target. A stride-4 detection head is added, strided convolution is replaced by lossless space-to-depth, and contrast enhancement is prepended (Table 1, Section 2.1).

The present work begins from that same recipe, combining a P2/4 detection branch with two modules proposed here: LearnableContrast (Section 4.1), which adapts gamma and gain per tile rather than fitting one setting for the whole image, and EdgeSPD (Section 4.2), a lossless space-to-depth downsampler gated by a Sobel prior. YOLO26 is adopted as the base for deployment reasons as well as architectural ones, since its NMS-free head means no threshold requires re-tuning on a new road or camera and latency per frame does not depend on how many candidates a frame produces. Where this work departs from the literature is the step taken next: before any performance figure is reported, the dataset and the trained model's failures are submitted to a systematic audit (Section 3), and that audit determines what remains in the architecture (Section 4). The audit contradicted the design it was intended to support. Measured against the annotations rather than against visual impression, the distress in this dataset is not small, because a linear crack is marked as an elongated region rather than as a hairline object, so the stride-4 detection level was removed while its features were retained in the neck.

The wider finding is methodological rather than architectural. The two modules were tested both inside the proposed architecture and transplanted onto the unmodified baseline, under success criteria fixed before the runs began, and the gains they showed on the split that selected their checkpoints did not survive evaluation on the split that entered no decision. The contributions are therefore stated as follows.

\begin{enumerate}
\def\labelenumi{\arabic{enumi}.}
\tightlist
\item
  A reusable data-first audit protocol: box-geometry statistics at the training resolution, a per-instance failure decomposition of a trained baseline, and a model-assisted label-completeness check (Section 3).
\item
  A counter-trend architectural result: on region-annotated imagery the P2/4 detection head is unnecessary, since retaining the P2 features while removing the detection level gains 2.8 test and 0.9 validation mAP50 while cutting epoch time by 7.4\% and removing 75\% of all anchors (Section 7.1).
\item
  A paired, recipe-matched ablation showing that module-level gains measured on the selecting split systematically vanish or reverse under validation evaluation, while the detection-scale finding keeps its sign on both (Section 7.1).
\item
  A quantified account of the performance ceiling on the crack class, which is bounded by annotation geometry rather than by capacity (Section 7.2).
\item
  Two modules reported with their limits: LearnableContrast (494 parameters), a detector-supervised analogue of CLAHE active at inference, and EdgeSPD (2 parameters over SPD-Conv, though its rearrangement costs 8.85M at scale l). Neither survives validation evaluation in isolation (Section 7.1).
\end{enumerate}

\section{Related Work}\label{related-work}

Three bodies of work bear on this study. The pavement-detection literature is organised around detection scale: YOLOv8-PD {[}8{]}, SRD-YOLO {[}9{]} and YOLOv11n-CDL {[}10{]} each add a stride-4 detection head on stated small-object or thin-crack grounds, while MFF-YOLO {[}11{]}, EMG-YOLO {[}12{]}, BsS-YOLO {[}13{]}, SDFC-YOLO {[}14{]} and YOLO-ERCD {[}15{]} follow the same recipe with different components (Table 1). The second body supplies the components those detectors borrow to preserve weak signal. SPD-Conv {[}16{]} replaces strided convolution with a lossless space-to-depth rearrangement so that no pixel is discarded when resolution is reduced. For contrast, IA-YOLO {[}17{]}, GDIP-YOLO {[}18{]} and ERUP-YOLO {[}19{]} learn image-processing parameters jointly with the detector but fit one setting per whole image, whereas CLAHE {[}20{]}, the classical choice for low-contrast pavement imagery, is local but follows a fixed rule. Area attention {[}5{]} and BiFPN-style skips are inherited here unmodified rather than proposed, and YOLO26 {[}7{]} supplies the base detector. The third body is methodological: error-decomposition tools such as TIDE {[}21{]} and the diagnostic taxonomy of {[}22{]} separate a detector's failures by cause, but the pavement literature reports aggregate metrics and rarely audits its data or its failure modes before choosing an architecture.

\subsection{Comparison of mechanisms}\label{comparison-of-mechanisms}

\begin{table*}[!t]
\centering
\footnotesize
\setlength{\tabcolsep}{4pt}
\caption{Related work summary. Contrast, downsampling and P2 give each method's treatment of low-contrast input, early feature decimation, and whether a stride-4 detection head is added.}
\label{tab:1}
\begin{tabularx}{\textwidth}{XcXXXXX}
\toprule
Method (venue, yr) & Base & Key additions & Contrast & Downsampling & P2 head & Dataset — reported Δ mAP50 \\
\midrule
YOLOv8-PD (Sci. Rep. 2024) & YOLOv8n & BOT transformer, LSKA attention, C2fGhost, shared light head & — & strided conv & no & RDD2022 — 0.706 (+1.4) \\
EMG-YOLO (Front. Neurorob. 2024) & YOLOv5 & global-context C3, decoupled head, MPDIoU & — & strided conv & no & RDD2022 — (+2.9); ~2× model size \\
SRD-YOLO (AICIT 2024) & YOLOv8 & RFAConv, 4-level scale fusion for small targets & — & strided conv & \textbf{yes} & pavement distress — (+2.7) \\
MFF-YOLO (NNICE 2024) & YOLOv8 & BoTNet backbone, BiFPN neck, small-target focus & — & strided conv & — & road crack, complex background \\
SDFC-YOLO [14] & YOLO & dynamic downsampling module, feature-fusion compensation & — & adaptive sampling positions & — & pavement distress \\
YOLOv11n-CDL (JCEM 2026) & YOLO11n & double-stage attention, P2 small-object path & — & strided conv & \textbf{yes} & fine cracks / micro-potholes \\
BsS-YOLO [13] & YOLO & enhanced multi-path fusion, attention & — & strided conv & no & road crack \\
YOLO-ERCD (Sensors 2026) & YOLOv10s & R-CBAM, CAGC gamma aug., VFNM noise mod. & random per-channel gamma, \textbf{train-time only} & strided conv & no & HKPC — 0.677 (+2.1); RDD2022 — (+0.4) \\
SPD-Conv (ECML-PKDD 2022) & generic & space-to-depth + non-strided conv & — & \textbf{lossless SPD} & — & COCO-style small-object gains \\
IA-YOLO (AAAI 2022) & YOLOv3 & CNN-PP predicting DIP filter params & \textbf{global} learned filters (incl. gamma) & strided conv & no & foggy/low-light VOC \\
GDIP (ICRA 2023) & YOLOv3 & gated parallel DIP filters & \textbf{global} learned filters & strided conv & no & adverse weather \\
ERUP-YOLO (WACV 2025) & YOLO & unified image-adaptive filters & \textbf{global} learned filters & strided conv & no & adverse weather \\
\textbf{YOLO26-RD (this work)} & \textbf{YOLO26} (NMS-free) & LearnableContrast (494 p), EdgeSPD (+2 p), A2C2f, BiFPN skips; \textbf{P2 head removed after data audit} & \textbf{local (tile-wise), learned, active at inference} & \textbf{edge-gated lossless SPD} & \textbf{audited, then removed} & this dataset — \textbf{val 0.737 vs 0.709 (+2.8)} vs recipe-matched YOLO26; test 0.787 vs 0.773 \\
\bottomrule
\end{tabularx}
\end{table*}
Table 1 places YOLO26-RD alongside those works, and four distinctions follow. Every pavement detector in the table adds machinery for small objects without first measuring whether its own annotations contain them, whereas the architecture proposed here follows from that measurement (Section 3.2). LearnableContrast is local, learned end-to-end from the detection loss, and active at inference, a combination none of the prior contrast methods provides, since the learned filters are global and YOLO-ERCD's CAGC is switched off after training. EdgeSPD keeps the pixel-preserving property of SPD-Conv and adds a gradient prior identifying where thin structure lies, which nothing in the original mechanism distinguishes. And YOLO26-RD is, to our knowledge, the first pavement detector built on an end-to-end NMS-free head, which makes prediction count and latency per frame deterministic, a practical rather than an accuracy property. One caution applies to the last column of Table 1: each gain there comes from a different dataset and protocol, so the figures situate the works but are not comparable with one another or with ours.

\section{Dataset and Audit}\label{dataset-and-audit}

Because the audit determines the architecture in Section 4, we present it first. Together with the visual characteristics of Section 3.1, it supplies every design decision in Section 4: the size distribution of the annotated instances (Section 3.2) determines the detection scales, and the error modes of a trained model (Section 3.2) determine where capacity is reallocated, closing with an assessment of label completeness (Section 3.2).

\subsection{Data provenance, annotation protocol and dataset characteristics}\label{data-provenance-annotation-protocol-and-dataset-characteristics}

The dataset is drawn from production road-condition surveys carried out on public roads in Thailand by Infraplus Co., Ltd., using the vehicle-mounted system of Section 1. The camera is installed perpendicular to the road surface on a rigid instrument rail above the roof, the arrangement shown in Figure 2. This is operational inspection data rather than a research collection, captured in the course of commercial pavement assessment, so it spans the pavement types, surface ages, illumination conditions and roadside settings encountered across the surveyed network.

\begin{figure*}[!t]
\centering
\includegraphics[width=1.0\textwidth]{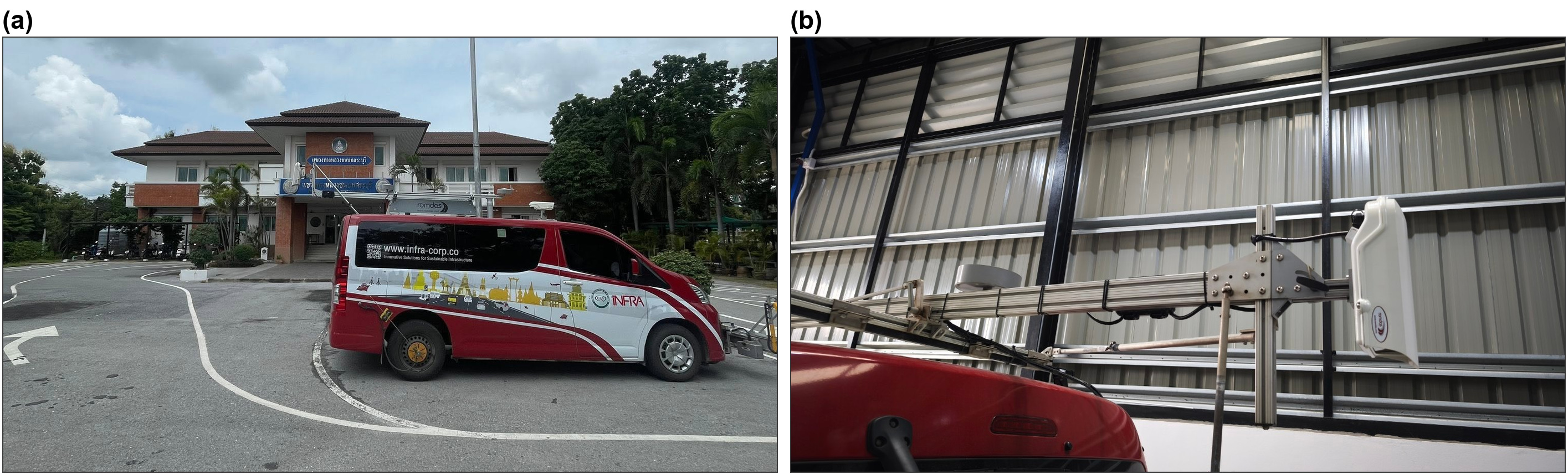}
\caption{Survey platform: \emph{(a)} the instrumented vehicle, and \emph{(b)} the roof-mounted instrument rail carrying the camera housing.}
\label{fig:2}
\end{figure*}
All annotation was performed in-house by Infraplus's pavement engineering team, whose members have more than five years of experience in pavement condition assessment, under a two-stage protocol: a labeller drew and classified every box, after which a second engineer reviewed the completed frame and either accepted it or returned it for correction, so every annotation reflects the judgement of at least two assessors. The criteria are those used in production inspection rather than criteria devised for this study, and they determine the box geometry the audit measures. An alligator crack is annotated as the region enclosing an interconnected network of fatigue cracking, taken as one instance rather than as its constituent cracks; a crack is a discrete linear fracture annotated with a box spanning its full visible length; and patching is the region covered by previous repair material. In every class the box encloses the full extent of the distress as it appears in the frame, which is why the annotations are large regions rather than pixel-thin segments, and Section 3.2 shows that consequence to be decisive for the architecture.

The imagery is grayscale content stored as RGB JPEG at high resolution, predominantly 2800×1800 and 2504×1600 px. The three classes are annotated across 7,618 images split 6,563 train / 504 test / 551 validation, carrying 9,194 instances (Table 2). Scenes are sparse, with a median of one instance per annotated frame, and frames containing no distress are retained in every split as hard negatives. Split integrity was verified before any measurement: zero degenerate boxes and zero source-frame leakage between splits, checked on de-hashed filename stems.

Figure 3 shows three representative frames with detections overlaid, one per distress class. Two visual characteristics of the imagery motivate the two modules of Section 4. The first is low and uneven contrast: the imagery is grayscale, so distress is separated from sound pavement by a few gray levels, and illumination varies within single frames, so one global correction must choose between brightening a shadowed region and preserving a sunlit one. What the data demands is a correction that is local, learned and applied at inference (Section 4.1). The second is thin structure against downsampling: frames are captured at about 2800 px width but trained at 640², so a crack a centimetre or two wide survives as only a few pixels and a cascade of stride-2 convolutions can discard it before any deep feature is built from it, which lossless but undirected space-to-depth only partly addresses (Section 4.2).

\begin{table}[!t]
\centering
\footnotesize
\setlength{\tabcolsep}{4pt}
\caption{Dataset splits: image counts, background frames, and per-class instance counts.}
\label{tab:2}
\begin{tabular}{lccccc}
\toprule
split & images & backgrounds & alligator & crack & patching \\
\midrule
train & 6,563 & 1,711 (26\%) & 1,742 & 5,380 & 662 \\
test & 504 & 39 (8\%) & 241 & 418 & 65 \\
val & 551 & 117 (21\%) & 237 & 385 & 64 \\
\bottomrule
\end{tabular}
\end{table}
\begin{figure*}[!t]
\centering
\includegraphics[width=0.92\textwidth]{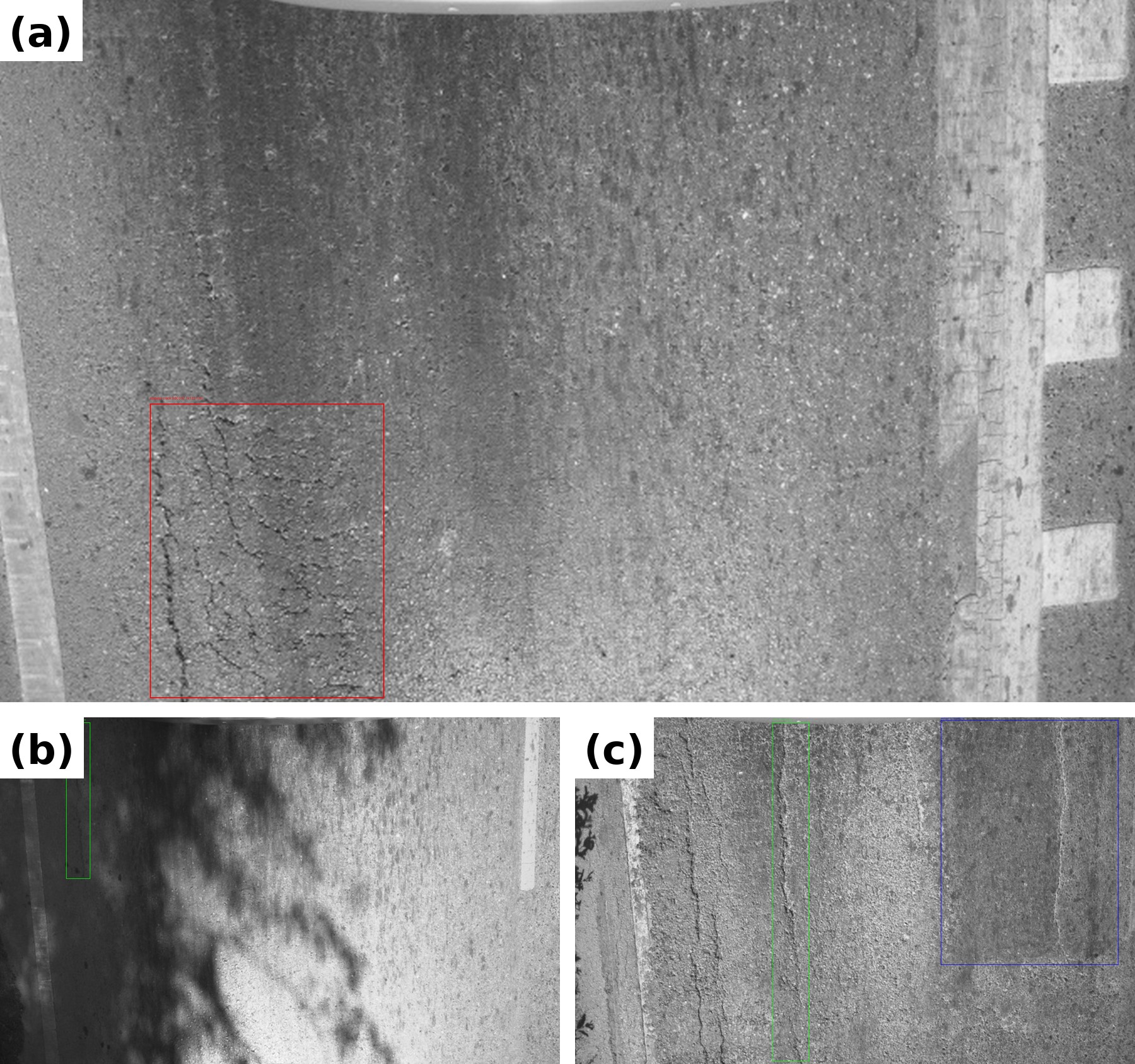}
\caption{Down-facing survey frames with YOLO26-RD detections overlaid, annotated by class, confidence and physical extent. \emph{(a)} An \texttt{alligator crack} region (red box). \emph{(b)} A \texttt{crack} under dappled tree shadow (green box). \emph{(c)} A \texttt{crack} and a \texttt{patching} region in one frame (green and blue boxes). Discussed in the text above.}
\label{fig:3}
\end{figure*}
\subsection{Audit: detection scales, failure modes and label completeness}\label{audit-detection-scales-failure-modes-and-label-completeness}

The recipe this paper questions assumes that pavement distress consists of small objects. Table A1 in Appendix A tests that assumption against the annotations themselves and against a trained baseline, and every row contradicts it. At the 640² training resolution only 1.28\% of instances are COCO-small and 70.37\% are large, because a linear crack is annotated as the region enclosing it: the median crack box is 138 × 1,476 px at aspect 9.5:1, spanning most of the frame height. A stride-4 detection level would nonetheless supply 75.3\% of all anchors, evaluating every one of them at each forward pass, so the conventional recipe spends three quarters of its assignment budget at a scale where 1\% of instances live.

The failure decomposition of the unmodified YOLO26-l reference points the same way. Across 724 instances the complete miss dominates every class, at 10\% of alligator, 16\% of crack and 34\% of patching ground truth, whereas localization error accounts for about 7\% of crack instances and none of the patching ones, and inter-class confusion for two instances in the whole split. The bottleneck is therefore sensitivity rather than box placement, which is what the freed anchor budget is directed at. Two further rows bound what any architecture can achieve here. Crack geometry shifts between train and test, the median crack spanning 0.90 of frame height on one and less on the other, so part of the residual error is distribution shift rather than model capacity. And a model-assisted review of the 113 confident false positives found most to be real but unlabelled distress, 74 of them cracks, so measured precision is a lower bound for every arm reported in this paper. The labels were deliberately left untouched, since correcting them mid-study would have placed the revision downstream of knowing which model each correction would favour.

\section{YOLO26-RD Architecture}\label{yolo26-rd-architecture}

YOLO26-RD is built on the YOLO26 template, inheriting its end-to-end one-to-one head with \texttt{reg\_max\ =\ 1} direct box regression and its C3k2, SPPF and C2PSA blocks, with inputs normalized to {[}0, 1{]}. We adopt this base unmodified for two reasons: NMS-free inference yields a deterministic prediction count and latency per frame, which matters for survey-vehicle deployment (Section 8), and using an off-the-shelf template makes the architectural comparisons of Sections 6 and 7 exact rather than approximate. The specification below is given at scale l, the scale released with this paper; the five scales differ only in the YAML \texttt{scales} entry, so the topology is common to all of them.

Four modifications distinguish YOLO26-RD from that template, and each is traceable to a specific measurement rather than to design intuition. We list them in the order the audit produced them, noting for each which measurement of Section 3 motivates it and, where applicable, which ablation arm in Section 7 tests it.

\begin{enumerate}
\def\labelenumi{\arabic{enumi}.}
\tightlist
\item
  A LearnableContrast stem at the input (layer 0, Section 4.1) answers the low-and-uneven-contrast characteristic of Section 3.1: it applies a local, per-tile gamma/gain correction learned from the detection loss and active at inference, where a global correction cannot serve shadowed and sunlit regions of one frame simultaneously.
\item
  EdgeSPD replaces the two early strided convolutions at P2→P3 and P3→P4 (Section 4.2), answering the thin-structure characteristic of the same section: downsampling becomes lossless, and a Sobel gradient prior amplifies edge-bearing locations before the rearrangement, so a 0.5--1.1 px crack line cannot be discarded before deep features are built from it. The gradient gate itself costs two parameters per instance, but the space-to-depth rearrangement it steers presents four times as many channels to the fusion convolution as the strided convolution it replaces, and Section 4.3 reports what that costs in absolute terms at scale l.
\item
  No detection head at stride 4. The stride-4 (P2/4) branch is retained in the neck and fused into the bottom-up path as features but carries no \texttt{Detect} level, the direct consequence of Section 3.2, where a P2 head would supply 75.3\% of all anchors at a scale holding essentially no eligible ground truth. Only Detect(P3, P4, P5) remains, and the freed anchor and assignment budget is what Section 3.2 asks to be spent on the miss rate. An ablation arm at scale s tests this decision (Section 7.1): restoring the head costs 0.028 selection-split and 0.009 validation mAP50 while adding 7.4\% to epoch time, and gains 0.007 validation mAP50-95.
\item
  Area attention and BiFPN-style skips are inherited from the YOLOv12/BiFPN lineage rather than derived from our audit, and we label them as such: A2C2f blocks at backbone-P4 and head-P3, and additional skips carrying the backbone P3/P4 outputs directly into the bottom-up path alongside the standard top-down/bottom-up PAN flow. Section 7.1 reports what enabling genuine attention at head-P3 actually costs. Figure 4 shows the resulting data flow together with the internals of the two proposed modules; Sections 4.1 and 4.2 define those modules formally, and Section 4.3 gives the layer-by-layer specification.
\end{enumerate}

\begin{figure*}[!t]
\centering
\includegraphics[width=0.92\textwidth]{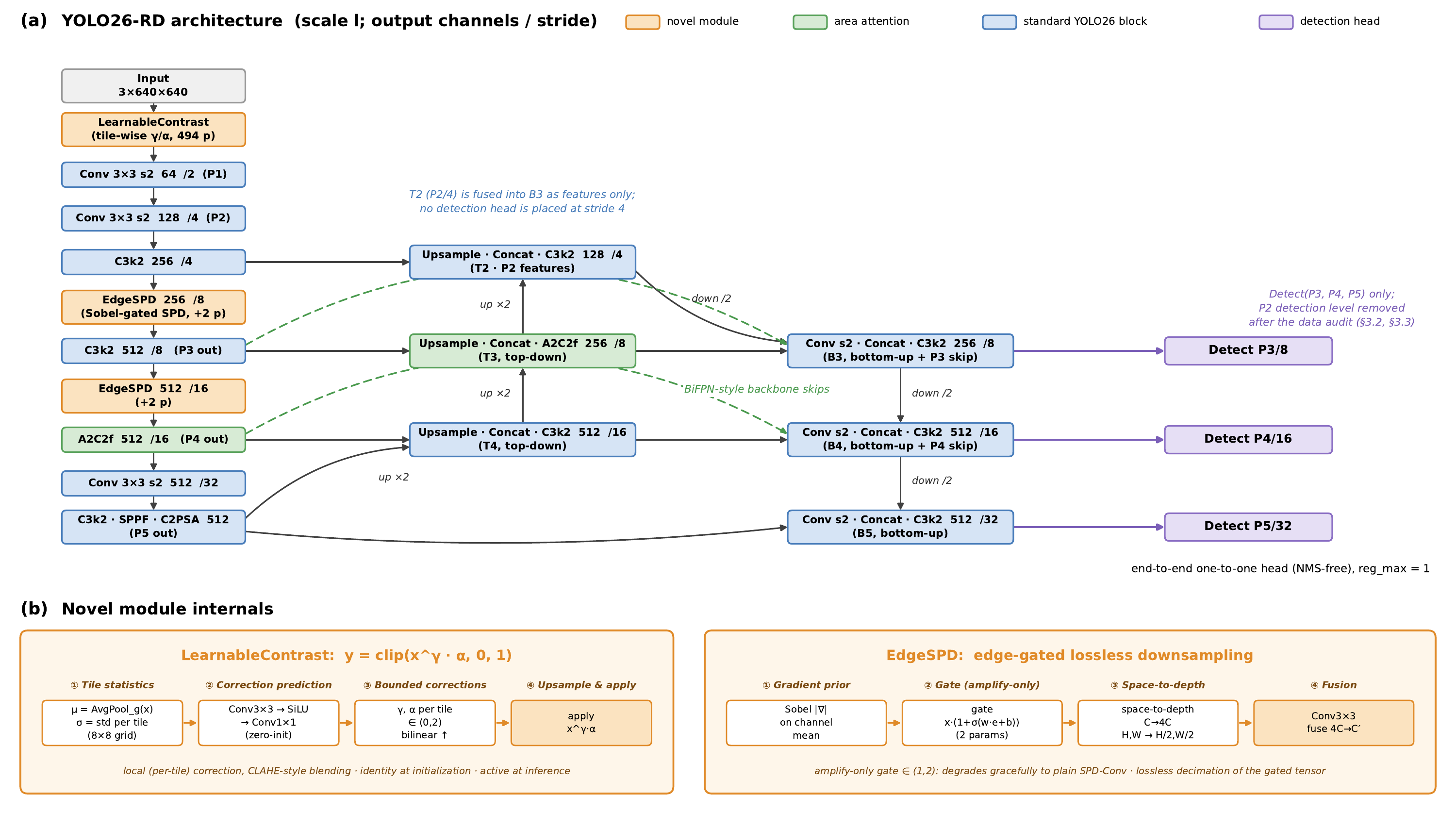}
\caption{YOLO26-RD topology. \emph{(a)} End-to-end data flow, blocks annotated with output channels and stride; orange marks the two proposed modules, green the area-attention blocks, and dashed green arcs the BiFPN-style backbone skips. \emph{(b)} Internals of LearnableContrast and EdgeSPD. Specified layer by layer in Table A2 and discussed in Sections 4.1 to 4.3.}
\label{fig:4}
\end{figure*}
\subsection{LearnableContrast: a differentiable local-contrast stem}\label{learnablecontrast-a-differentiable-local-contrast-stem}

Section 3.1 established what the data requires of a contrast correction: that it be local, learned from the detection loss, and active at inference. The classical method satisfying the first of those requirements is CLAHE (Contrast Limited Adaptive Histogram Equalization), which divides the image into tiles, equalizes each tile independently, then blends across tile boundaries so the result is not blocky. CLAHE is not learned, however: it is a fixed, hand-tuned preprocessing rule applied identically regardless of what the downstream detector needs from the image. LearnableContrast retains CLAHE's structure, per-tile correction with cross-tile blending, while replacing the fixed histogram-equalization rule with a small convolutional network trained end-to-end with the detector, so the correction it learns is the one that reduces detection loss rather than the one that appears visually equalized.

Figure 5 summarizes this data flow, separating the identity path from the correction path; the formal definition is given in Appendix B.

\begin{figure*}[!t]
\centering
\includegraphics[width=0.98\textwidth]{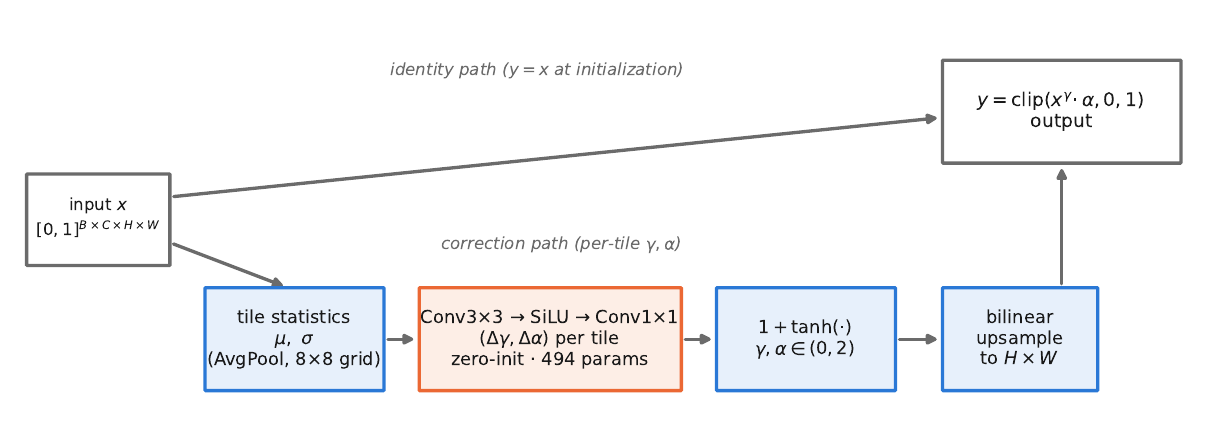}
\caption{LearnableContrast data flow. The identity path (top) carries the image unchanged to the output rule, while the correction path (bottom) reduces it to per-tile statistics, predicts the per-tile corrections with the module's only learnable block (orange, 494 parameters) and upsamples them before the pixelwise application. Blue blocks have no learnable parameters. Discussed in the text above.}
\label{fig:5}
\end{figure*}
Each stage has a specific role. Tile statistics reduce the image to an 8×8 grid of per-tile mean brightness and spread at no parameter cost. A 3×3 convolution mixes those statistics across neighbouring tiles so corrections stay spatially coherent, and a SiLU and a 1×1 convolution reduce them to two values per tile, Δγ and Δα; because this network sees only the statistics and never the raw image, it can condition on local brightness and contrast but cannot memorise image content, and almost the whole 494-parameter budget lies in these two convolutions. A tanh bounds the result to γ, α ∈ (0, 2), placing the identity at the centre of the range, and bilinear upsampling back to full resolution supplies the cross-tile blending that prevents visible blocking before y = clip(x\^{}γ · α, 0, 1) is applied pixelwise. The final 1×1 convolution is zero-initialised, so the module is exactly the identity on the first forward pass and cannot degrade early training; every later departure is a correction the detection loss selected. Three properties distinguish it from prior work (Section 2.1): the correction is local rather than one setting per image as in IA-YOLO, GDIP and ERUP-YOLO; it is learned end-to-end from the detection loss rather than from a fixed rule as in CLAHE; and it stays active at inference, unlike YOLO-ERCD's CAGC.

\subsection{EdgeSPD: edge-gated lossless downsampling}\label{edgespd-edge-gated-lossless-downsampling}

Section 3.1 established the requirement here: downsampling that is lossless and informed about where thin structure lies. SPD-Conv {[}16{]} supplies the first half: instead of discarding pixels, it rearranges the four pixels of every 2×2 block into four separate channels (a space-to-depth operation), so every input pixel survives into the next layer, and a 3×3 convolution then fuses the resulting 4C channels back down to a chosen output width. Nothing is discarded, and the trade is spatial resolution for channel depth rather than information for compute. Its limitation is that it supplies no second half: it treats every pixel identically, so a flat patch of pavement and a crack edge are rearranged and fused by exactly the same operation, with no signal telling the fusion convolution which locations matter more. EdgeSPD adds that signal, computed before the rearrangement, at a cost of two parameters.

Figure 6 summarizes this data flow, separating the edge-gate path from the lossless rearrangement it modulates; the formal definition is given in Appendix B.

\begin{figure*}[!t]
\centering
\includegraphics[width=0.98\textwidth]{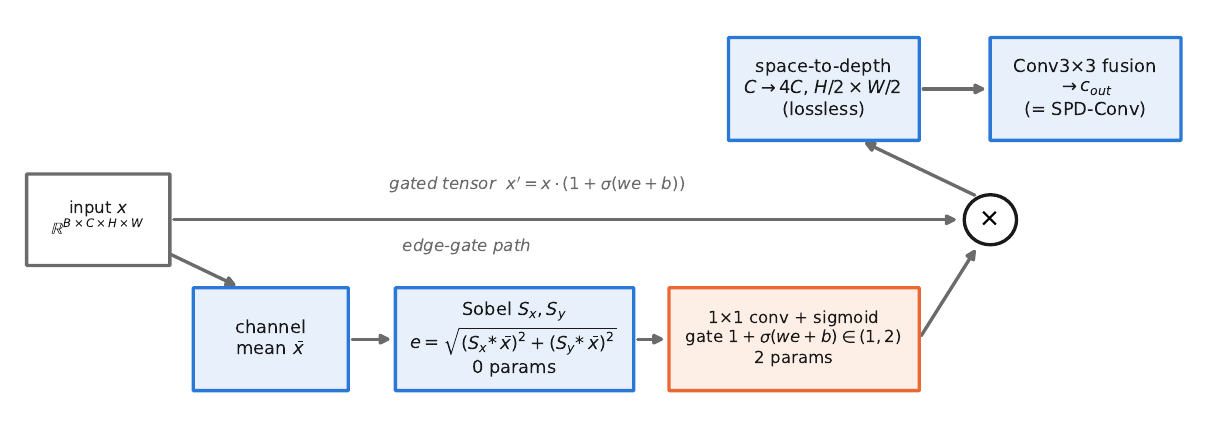}
\caption{EdgeSPD data flow. The edge-gate path (bottom) computes a Sobel gradient-magnitude map from the channel mean, with no learnable parameters, and maps it through the module's entire learnable budget of 2 parameters (orange) into a multiplicative gate. The gate amplifies edge-bearing locations before the lossless space-to-depth rearrangement and 3×3 fusion (top right). Discussed in the text above.}
\label{fig:6}
\end{figure*}
Each stage has a specific role. The gradient prior convolves the channel-mean of the incoming map with a fixed 3×3 Sobel pair, responding to the sharp, thin brightness transitions a crack produces rather than to slow illumination changes. The gate can only scale features upward, by a factor strictly in (1, 2), so it emphasises edge-bearing locations without suppressing anything, and it is initialised so that the module begins as plain SPD-Conv and departs from it only if the detection loss rewards doing so. Space-to-depth then rearranges the gated map losslessly into four sub-maps concatenated along channels, so no pixel is discarded, and a fusion convolution reduces the result to the target width. The learnable cost is two parameters per instance, but the rearrangement presents four times as many channels to that fusion convolution as the strided convolution it replaces, which costs 8.85M parameters at scale l and 2.21M at scale s (Section 4.3); the two-parameter figure is the gate's cost relative to SPD-Conv, not the cost of adopting lossless downsampling.

\subsection{Layer-by-layer specification}\label{layer-by-layer-specification}

Table A2 in Appendix A specifies the complete network layer by layer at scale l, giving per-layer parameter counts, output widths, strides and roles. Two settings in it are conditioned on scale and inherited from the base template rather than chosen here: C3k inner blocks replace plain bottlenecks at scales m, l and x, and the A2C2f blocks gain a residual branch and an expansion of 1.2 at l and x. The resulting scale-l network carries 36.22M parameters unfused and 34.78M fused, exposes 8,400 anchors at 640² input, and runs its inference stage in 2.64 ms per image as a TensorRT FP16 engine at batch one, against 12.42M and 1.64 ms at scale s.

Removing the stride-4 detection level does not reduce capacity. Layer 20 retains the P2/4 branch and feeds the bottom-up path, so high-resolution detail still reaches the P3 level, and the parameter count in fact rises against a P2 to P5 variant, 36.22M against 34.85M at scale l, because YOLO26 ties the classification-branch width to the narrowest detection level present. The change is a reallocation of assignment and loss budget toward the scales where objects exist. The two modules differ in cost by orders of magnitude and must be quoted separately: the two EdgeSPD layers hold 11.80M parameters, 32.6\% of the network, of which 8.85M at scale l and 2.21M at scale s is the price of lossless downsampling over the strided convolutions it replaces, whereas LearnableContrast is near-free at 494 parameters at every scale.

\section{Experimental Setup}\label{experimental-setup}

All arms in this paper share a single training configuration, summarized in Table 3. Fixing every hyperparameter across arms is what allows the differences reported in Sections 6 and 7 to be attributed to architecture: an arm that changes a training setting rather than a structural one is not comparable to the others and none is reported here.

\begin{table*}[!t]
\centering
\small
\setlength{\tabcolsep}{6pt}
\caption{Training and evaluation configuration, identical for every arm.}
\label{tab:3}
\begin{tabularx}{\textwidth}{lX}
\toprule
item & setting \\
\midrule
Hardware & 1 × NVIDIA RTX 5090 (32 GB); 2 × RTX 5090 for the scale-x arms, which do not fit one card at batch 32 \\
Framework & PyTorch 2.12 / CUDA 13; Ultralytics 8.4.115 fork containing both modules \\
Initialization & from scratch; no pretrained weights in any arm \\
Input resolution & 640 × 640 \\
Batch & 32, with gradient accumulation to a nominal batch of 64 \\
Epochs & 150 \\
Optimizer & MuSGD, lr 0.01 (framework auto-selection), cosine schedule, AMP \\
Augmentation & mosaic 0.5 (\texttt{close\_mosaic} 30), flipud 0.5, fliplr 0.5; no rotation, mixup or cutmix \\
Seed & 0 \\
Checkpoint selection & fitness = mAP50-95 on the 504-image test split (weights [P, R, mAP50, mAP50-95] = [0, 0, 0, 1]) \\
Final evaluation & 551-image validation split, evaluated once per model after all decisions were frozen \\
\bottomrule
\end{tabularx}
\end{table*}
Every arm was trained from scratch. The LearnableContrast stem shifts every subsequent layer index and EdgeSPD replaces the strided convolutions, so no off-the-shelf YOLO26 checkpoint transfers into the proposed architecture by name and shape, and warm-starting only the baseline would have confounded architecture with initialization. Two settings in Table 3 were fixed by measurement rather than convention: mosaic was reduced to 0.5 because tiling truncates near-full-frame crack boxes and injects label noise, and batch size was fixed at 32 because 32, 48 and 64 gave the same epoch time at full GPU utilization. Accuracy is reported as mean average precision. mAP50 applies a single intersection-over-union threshold of 0.50, so it measures whether a distress was found and roughly located, while mAP50-95 averages thresholds from 0.50 to 0.95 and so additionally rewards precise box placement. The two splits play different roles, and their names are the reverse of the usual convention, so the roles rather than the names should be read: the 504-image split, called the test split in Table 2, selected every checkpoint and therefore carries selection bias, whereas the 551-image validation split was evaluated once after all decisions were frozen and is what the headline claims rest on. Every figure below names the split it was measured on. Every arm ran the same 150-epoch schedule, so the scale-l comparison of Section 6, which pairs YOLO26-RD-l at 36.22M parameters against the YOLO26-l reference at 24.75M, is recipe-matched, as is every ablation arm in Table 5 and every member of the two reference families.

\section{Comparison with the Base Model at Scale l}\label{comparison-with-the-base-model-at-scale-l}

\begin{table*}[!t]
\centering
\small
\setlength{\tabcolsep}{6pt}
\caption{YOLO26-RD-l against the YOLO26-l reference on the test split.}
\label{tab:4}
\begin{tabular}{lcccc}
\toprule
 & baseline: YOLO26-l & \textbf{YOLO26-RD-l} & absolute & relative \\
\midrule
\textbf{mAP50} & 0.778 & \textbf{0.809} & +0.031 & +3.98\% \\
\textbf{mAP50-95} & 0.467 & \textbf{0.497} & +0.030 & +6.42\% \\
precision & \textbf{0.857} & 0.830 & −0.027 & −3.15\% \\
recall & 0.706 & \textbf{0.717} & +0.011 & +1.56\% \\
AP50, alligator & 0.821 & \textbf{0.854} & +0.033 & +4.02\% \\
AP50, crack & 0.721 & \textbf{0.744} & +0.023 & +3.19\% \\
AP50, patching & 0.792 & \textbf{0.828} & +0.036 & +4.55\% \\
AP50-95, alligator & 0.457 & \textbf{0.473} & +0.016 & +3.50\% \\
AP50-95, crack & 0.359 & \textbf{0.385} & +0.026 & +7.14\% \\
AP50-95, patching & 0.585 & \textbf{0.634} & +0.049 & +8.38\% \\
parameters (fused) & \textbf{24.75M} & 34.78M & +10.03M & +40.5\% \\
anchors 640² & 8,400 & 8,400 & — & — \\
epochs / s per epoch & 150 / \textbf{45.4} & 150 / 112.1 & — & — \\
TensorRT FP16 ms/img & \textbf{2.02} & 2.64 & +0.62 & +30.7\% \\
\bottomrule
\end{tabular}
\end{table*}
Table 4 sets the two arms side by side, both trained from scratch on identical data, splits and seed. On the 504-image test split, which selected both checkpoints and therefore carries selection bias, YOLO26-RD-l leads the YOLO26-l reference at 0.809 mAP50 against 0.778 and 0.497 mAP50-95 against 0.467, and it leads in all six per-class entries. The claims of this section therefore rest on the 551-image validation split, which entered no decision, and there the result changes. The mAP50 margin collapses to 0.738 against 0.734, a fifth of the ±0.015 resolution limit of Section 7.1, so the two architectures find essentially the same instances, and the per-class figures make that literal, with crack identical at 0.650. What survives is the stricter metric: validation mAP50-95 is 0.474 against 0.461, a margin of 0.013 whose sign is uniform across all three classes, with precision rising from 0.805 to 0.822 while recall falls from 0.681 to 0.665. The gain at scale l is therefore localization rather than sensitivity, the proposed model drawing tighter boxes around instances both models already find, whereas at scale s the same architectural change instead buys sensitivity, raising recall from 0.686 to 0.748. That improvement is not free. YOLO26-RD-l carries 34.78M fused parameters against 24.75M, an increase of 40.5\%, and spends 2.64 ms against 2.02 ms in the compiled inference stage at batch one, an increase of 30.7\%. A practitioner at this capacity therefore buys 0.013 validation mAP50-95, and nothing measurable on validation mAP50, in exchange for those costs, which is a defensible trade only where boundary precision is worth more than throughput.

\section{Ablation and Comparative Study}\label{ablation-and-comparative-study}

This section separates two questions that the same set of runs answers, and gives each its own table. Table 5 is the module ablation: it holds capacity and recipe fixed at scale l, the capacity at which the released model is deployed, and asks which of the design decisions in Section 4 carries the margin reported in Section 6. Table 6 is the benchmark comparison: it places the proposed architecture, the unmodified base model and an external detector family side by side across their full scale ranges, so that the proposed model is judged against what a practitioner would otherwise train rather than against its own ablations alone. Table 5 reports the selection split alone, which bounds what it can establish, while Table 6 presents the selection-split scale comparison plotted in Figures 7 and 8; validation scale results are reported in Sections 7.2 and 7.3. Section 7.1 shows why the two splits must still be interpreted separately and why a claim stated on the selection split alone is not evidence of generalization.

Within Table 5 each arm answers one question about the two proposed modules at scale l, the capacity compared in Section 6. The grid varies LearnableContrast between present and absent and EdgeSPD between zero, one and two locations, and a disabled EdgeSPD location is replaced by a parameter-matched plain space-to-depth block rather than by a strided convolution, so every contrast within the grid isolates the Sobel gate and the number of gated downsamples rather than the choice of downsampling operator. The four intermediate arms and the proposed model were trained from scratch under the configuration of Table 3, for 150 epochs and at the nominal optimizer batch of 64. The baseline row is the YOLO26-l reference of Table 4, which ran the same schedule at 45.4 s per epoch against the proposed model's 112.1 s, so every row of the grid is recipe-matched and the contrasts differ only in architecture.

\begin{table*}[!t]
\centering
\scriptsize
\setlength{\tabcolsep}{3pt}
\caption{Module ablation of YOLO26-RD-l on the 504-image test split. LC is the LearnableContrast stem; EdgeSPD gives the number of gated downsampling locations, disabled locations being replaced by parameter-matched plain space-to-depth. The baseline row is the unmodified YOLO26-l of Table 4 and is outside the proposed graph. No arm was evaluated on the validation split (Section 7.1). Bold marks the best value per column.}
\label{tab:5}
\begin{tabular}{lcccccccccc}
\toprule
\# & arm & LC & EdgeSPD & mAP50 & mAP50-95 & P & R & AP50 alligator & AP50 crack & AP50 patching \\
\midrule
— & YOLO26-l baseline (not the RD graph) & no & — & 0.778 & 0.467 & \textbf{0.857} & 0.706 & 0.821 & 0.721 & 0.792 \\
B1 & LearnableContrast only & yes & 0 & 0.782 & 0.487 & 0.787 & \textbf{0.744} & 0.805 & 0.737 & 0.804 \\
B2 & EdgeSPD, one location & no & 1 & 0.790 & 0.481 & 0.838 & 0.714 & 0.839 & 0.742 & 0.790 \\
B3 & EdgeSPD, two locations & no & 2 & 0.794 & 0.494 & 0.854 & 0.684 & 0.842 & 0.735 & 0.806 \\
B4 & LearnableContrast + EdgeSPD, one location & yes & 1 & 0.786 & 0.493 & 0.813 & 0.708 & 0.845 & 0.727 & 0.787 \\
— & \textbf{YOLO26-RD-l} (proposed, Table 4) & yes & 2 & \textbf{0.809} & \textbf{0.497} & 0.830 & 0.717 & \textbf{0.854} & \textbf{0.744} & \textbf{0.828} \\
\bottomrule
\end{tabular}
\end{table*}
Every arm in the grid exceeds the YOLO26-l baseline on both aggregate metrics, so each module improves the base architecture on its own. The contrast stem alone raises mAP50 from 0.778 to 0.782 and mAP50-95 from 0.467 to 0.487. Gated downsampling reaches 0.790 and 0.481 at one location, and 0.794 and 0.494 at two. The best model of the six combines them. With the contrast stem and both gated downsampling locations, three module insertions in total, YOLO26-RD-l returns 0.809 mAP50 and 0.497 mAP50-95, and it leads on all three per-class AP50 entries: 0.854 for alligator, 0.744 for crack and 0.828 for patching. The two modules also divide the operating point as their premises predict. The contrast stem alone gives the highest recall of any arm, 0.744, and the lowest precision, 0.787. Gated downsampling without the stem gives the lowest recall, 0.684, at a precision of 0.854 that matches the baseline. The full configuration sits between them, at 0.717 recall and 0.830 precision.

Read as a factorial, however, the grid does not decompose into two independent contributions. Against the baseline the contrast stem adds 0.004 mAP50 and gated downsampling at two locations adds 0.016, but the two together add 0.031. The joint effect therefore exceeds the sum of the separate effects by 0.011. On mAP50-95 the corresponding figures are 0.020, 0.027 and 0.030, so there the joint effect falls 0.017 below the sum instead. The paired contrasts locate the interaction. The second gated location is worth 0.004 mAP50 without the contrast stem (B2 to B3) and 0.023 with it (B4 to the proposed model). The stem is worth −0.004 at one gated location (B2 to B4) and +0.015 at two (B3 to the proposed model), so the sign of its effect changes with the other module's setting. Neither module has a single effect size on these data, and one run per cell cannot separate the interaction from noise.

\begin{table}[!t]
\centering
\footnotesize
\setlength{\tabcolsep}{4pt}
\caption{Benchmark comparison across scale ranges. All fifteen arms trained from scratch under the 150-epoch recipe of Table 3 on the same dataset, splits and resolution, each evaluated at its fitness-selected checkpoint on the 504-image test split. Bold marks the best value per column within each family.}
\label{tab:6}
\begin{tabular}{lccc}
\toprule
model & params & mAP50 & mAP50-95 \\
\midrule
\multicolumn{4}{l}{\textbf{YOLO26-RD (proposed)}} \\
YOLO26-RD-n & 3.13M & 0.757 & 0.460 \\
YOLO26-RD-s & 12.42M & 0.787 & 0.469 \\
YOLO26-RD-m & 31.61M & 0.794 & 0.488 \\
\textbf{YOLO26-RD-l} (Table 4) & 36.22M & \textbf{0.809} & \textbf{0.497} \\
YOLO26-RD-x & 81.36M & 0.801 & 0.495 \\
\multicolumn{4}{l}{\textbf{YOLO26 (base model, recipe-matched)}} \\
YOLO26-n & 2.38M & 0.773 & 0.467 \\
YOLO26-s & 10.01M & 0.773 & 0.469 \\
YOLO26-m & 20.35M & 0.777 & 0.474 \\
\textbf{YOLO26-l} (Table 4 baseline) & 24.75M & \textbf{0.778} & 0.467 \\
YOLO26-x & 58.82M & \textbf{0.778} & \textbf{0.479} \\
\multicolumn{4}{l}{\textbf{YOLOv12 (external family, recipe-matched)}} \\
YOLOv12-n & 2.56M & 0.679 & 0.445 \\
YOLOv12-s & 9.23M & 0.752 & 0.463 \\
YOLOv12-m & 20.11M & 0.772 & 0.471 \\
YOLOv12-l & 26.34M & 0.782 & \textbf{0.482} \\
YOLOv12-x & 59.05M & \textbf{0.792} & 0.481 \\
\bottomrule
\end{tabular}
\end{table}
Table 6 is read in two places: Section 7.2 takes the proposed family's own capacity response, and Section 7.3 takes the two reference families and the comparisons between them, while Figure 7 plots its selection-split columns against parameter count. Before either reading, the parameter column bears on how the rows may be compared at all, because a shared scale letter does not denote a shared capacity. The proposed family carries 24\% to 55\% more parameters than YOLO26 at the same letter and 22\% to 57\% more than YOLOv12, the excess arising chiefly from the space-to-depth rearrangement inside EdgeSPD, which costs 8.85M at scale l, together with the retained stride-4 neck branch and the 494-parameter contrast stem, so a row-wise comparison at matched scale credits the proposed architecture with capacity as well as with design. Table 6 answers that objection within itself, by permitting arms of unequal scale but comparable or unfavourable capacity to be set against one another. YOLO26-RD-l returns 0.809 mAP50 and 0.497 mAP50-95 at 36.22M parameters against 0.778 and 0.479 for YOLO26-x at 58.82M and 0.792 and 0.481 for YOLOv12-x at 59.05M, so it exceeds the largest arm of both reference families on both metrics while carrying about 61\% of their parameters, and all four of those margins, 0.031 and 0.018 against the base model and 0.017 and 0.016 against the external family, lie outside the ±0.015 resolution limit of Section 7.1. One scale down the same comparison weakens to parity rather than reversing, YOLO26-RD-m returning 0.794 and 0.488 at 31.61M against the two scale-x arms at roughly 54\% of their parameters, with three of its four margins inside the resolution limit; one scale further down it fails, since YOLO26-RD-s trails both scale-x arms on mAP50-95 at 0.469. The lead reported in Section 7.3 is therefore not an artefact of the matched-scale parameter overhead from scale m upward, since the proposed model retains it against references holding half again as many parameters, and the counts quoted for the proposed family are the unfused ones, 36.22M at scale l against the 34.78M fused figure of Table 4, so these ratios use the larger of the two available counts for it. The base model's own column supplies the contrast that governs the shape of Figure 7, since YOLO26 spans only 0.773 to 0.778 mAP50 from 2.38M to 58.82M parameters and so converts almost none of the capacity it is given, whereas the proposed family converts the same range into 0.757 to 0.809.

\subsection{Module ablation and the transfer of measured gains}\label{module-ablation-and-the-transfer-of-measured-gains}

Table 5 varies the two modules inside the proposed graph at scale l, and neither proves redundant: removing LearnableContrast costs 0.015 mAP50 and removing the second EdgeSPD location costs 0.023, so no single-module arm reaches the proposed model's 0.809 mAP50 and 0.497 mAP50-95. The two modules move the operating point in opposite directions, LearnableContrast alone giving the highest recall of any arm at 0.744 and the lowest precision at 0.787, and gated downsampling without the stem giving the lowest recall at 0.684 and a precision matching the baseline. This is the division of labour the audit anticipated. Three features bound those readings. The grid omits the cell with neither module, so deltas against the baseline also carry the graph changes separating the two architectures. It was scored on one split only, and gains measured there do not reliably survive: of four predictions fixed before their runs and scorable against both splits, three fail on validation data, EdgeSPD's isolated gain vanishing and the contrast stem reversing sign. And at 724 selection and 686 validation instances this dataset cannot resolve differences below roughly ±0.015 mAP50, a band estimated from the spread among recipe-identical arms, which four of the six adjacent-cell contrasts fall below. The transferable point concerns method rather than either module: an ablation scored only on the selecting split would have reported both transplanted modules as successes, whereas validation evaluation supports neither.

\subsection{Capacity response and the crack-class ceiling}\label{capacity-response-and-the-crack-class-ceiling}

YOLO26-RD was trained at five scales under the identical recipe, varying only the YAML \texttt{scales} entry, so the series isolates capacity from architecture. On the selection split both aggregate metrics peak at scale l, at 0.809 mAP50 and 0.497 mAP50-95, and fall marginally at x to 0.801 and 0.495, decrements of 0.008 and 0.002 that lie well inside the ±0.015 resolution limit of Section 7.1 and therefore establish parity rather than a decline. Validation evaluation inverts that ordering at the top: scale x returns 0.761 mAP50, the highest of any arm in the study, while l's selection-split lead collapses to parity with s and m at 0.738 against 0.737 and 0.735 (Table 6). What survives is the simpler statement that capacity helps, roughly monotonically, through the largest scale trained, which makes scale the one architectural axis here whose positive effect reproduces on data used in no decision.

Neither capacity nor module composition moves the bottleneck class. Validation AP50 for crack spans 0.650 to 0.681 across every arm evaluated on that split, and the cause is geometric rather than architectural. On the same split the proposed model's ratio of AP50-95 to AP50 is 0.80 for patching and 0.65 for alligator but 0.44 for crack, the signature of instances that are found but loosely bounded. For a median crack box of 33.6 × 345.8 px at 640² input, aspect 10.3:1, overlap is governed almost entirely by the short axis: a 10 px error in width drops IoU to 0.771, whereas the same error in length costs almost nothing at 0.972. The crack line itself is only 0.5 to 1.1 px wide at that resolution and the box edge around it is annotator-chosen, so the regression target is ambiguous in precisely the dimension that determines the score. Oriented boxes are no remedy, since 94.5\% of crack boxes are already elongated at least 3:1 along the image axes. Combined with the geometry shift of Section 3.2 and the label incompleteness of Section 3.2, this locates the crack ceiling in the data.
\#\# Reference baselines: YOLO26 scale family and YOLOv12

To situate the proposed architecture across its capacity range, YOLO26 was trained from scratch at scales n, m, l and x and YOLOv12 at all five scales, on the same dataset, splits and resolution, giving the three families of Table 6 and Figure 7. All three families ran the 150-epoch recipe of Section 5 on identical data, splits and resolution inside the same Ultralytics fork, so the differences that remain between them are architectural. Every one of the fifteen runs completed its schedule and was scored at its fitness-selected checkpoint on both splits, so no entry in the comparison is provisional and none substitutes a figure from one split for the other.

Figure 7 plots both metrics against parameter count on a logarithmic axis at the fitness-selected checkpoint on the selection split, and the three families exhibit distinct capacity behaviours. The unmodified YOLO26 is almost insensitive to scale, spanning 0.773 to 0.778 mAP50 across a 25-fold parameter range, an increment inside the resolution limit, and ordering non-monotonically on the stricter metric. YOLOv12 lies at the opposite extreme, spanning 0.679 to 0.792 mAP50 over the same range, so its accuracy depends strongly on parameter count and degrades sharply as capacity is reduced. YOLO26-RD lies between the two, rising from 0.757 mAP50 at 3.13M parameters to 0.809 at 36.22M.

Because the three curves differ in slope they intersect, but the proposed model's curve crosses each reference exactly once and does so between scales n and s in both panels, so the preferred architecture depends on the capacity budget only at the bottom of the ladder. The two reference curves cross each other near 26M parameters in panel (a), which is a comparison between the references rather than one this paper makes. At the smallest capacity trained the unmodified YOLO26 is the most accurate of the three, returning 0.773 mAP50 at 2.38M parameters against 0.757 for the proposed model at 3.13M and 0.679 for YOLOv12 at 2.56M. From scale s upward the ordering is fixed: YOLO26-RD returns the higher mAP50 against both references at every remaining scale, leading YOLO26 by 0.014 to 0.031 and YOLOv12 by 0.009 to 0.035, and from scale m upward it leads on both aggregate metrics simultaneously. The three larger scales are therefore the substantive result of this comparison, and they are unanimous. At m, l and x the proposed model wins all twelve pairwise contests against the two reference families, leading YOLO26 by 0.017, 0.031 and 0.023 mAP50 and by 0.014, 0.030 and 0.016 mAP50-95, and leading YOLOv12 by 0.022, 0.027 and 0.009 mAP50 and by 0.017, 0.015 and 0.014 mAP50-95. Eight of those twelve margins exceed the ±0.015 resolution limit of Section 7.1 and a ninth sits on it, so the lead is individually measurable on both metrics against YOLOv12 at scale m and against YOLO26 at scales l and x, and against YOLOv12 at scale l on mAP50 with its mAP50-95 margin of 0.015 falling exactly on the limit. The three margins below the limit, mAP50-95 against YOLO26 at scale m and both metrics against YOLOv12 at scale x, establish parity when read alone and take their weight from the uniformity of sign across the set rather than from their own size. The twelve contests are not independent, since they share three scales, two metrics and a single evaluation split, so no formal significance follows from their unanimity and none is claimed; what the pattern does exclude is the reading that the three sub-limit margins arise from noise about a common mean, because noise of that kind would be expected to place at least one of the twelve in the opposite direction. Averaged over the five scales the same ordering holds for the families as wholes, the proposed one returning 0.790 mAP50 and 0.482 mAP50-95 against 0.776 and 0.471 for YOLO26 and 0.755 and 0.468 for YOLOv12, margins of 0.014 and 0.011 over the base model and 0.034 and 0.013 over the external family, and the single most accurate model of the fifteen is YOLO26-RD-l on both metrics at 0.809 and 0.497. The proposed architecture consequently holds the accuracy lead over both reference families across the whole upper part of the ladder, from scale s on mAP50 and from scale m on both metrics, spanning the proposed family from 12.42M to 81.36M parameters and including the capacity at which the released model is deployed, and the only ordering that reverses anywhere is the one against the base model at the smallest capacity. These are selection-split values, and the validation ordering reported below does not follow them.

\begin{figure*}[!t]
\centering
\includegraphics[width=0.8\textwidth]{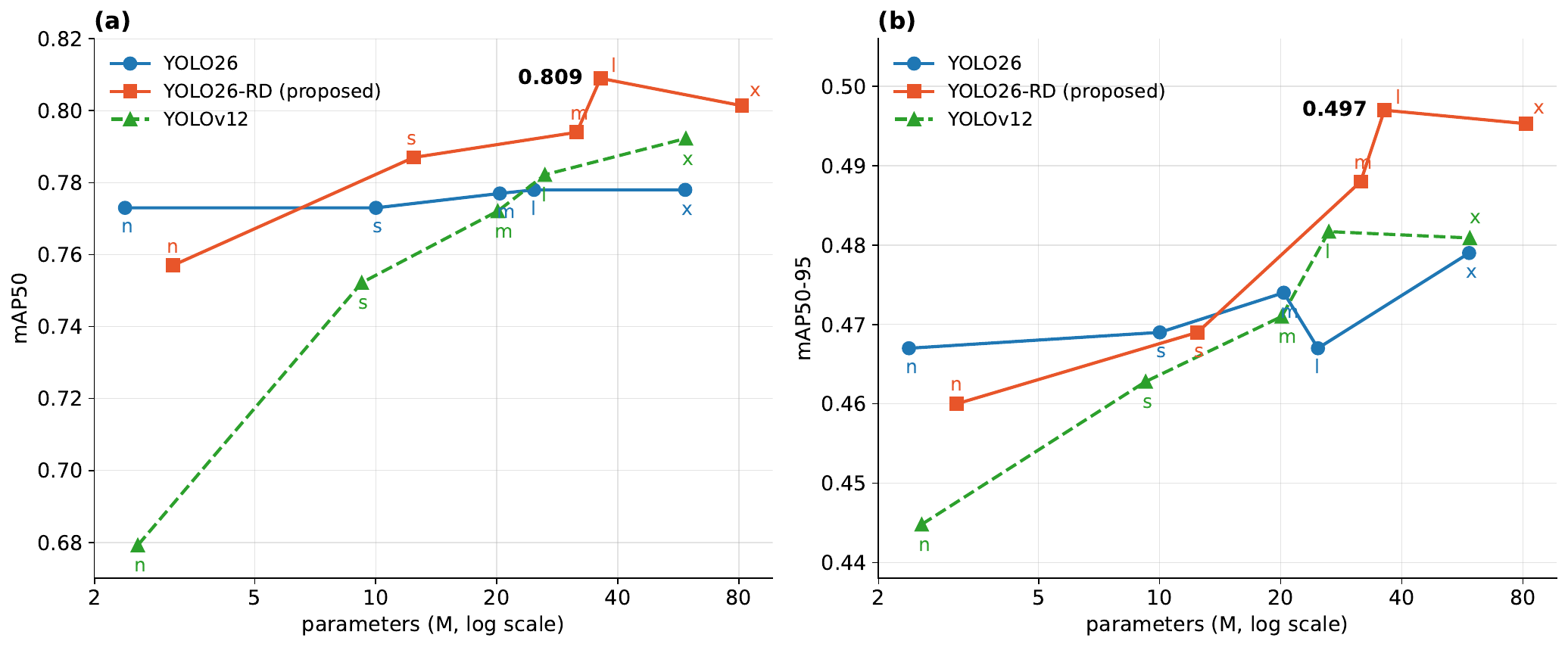}
\caption{Scale comparison of YOLO26 (blue circles), YOLO26-RD (orange squares) and YOLOv12 (green triangles, dashed) against parameter count on a logarithmic axis, scales n/s/m/l/x per family, at the fitness-selected \texttt{best.pt} checkpoint of every run. \emph{(a)} mAP50. \emph{(b)} mAP50-95. All points are selection-split values at completed runs. Values are tabulated in Table 6.}
\label{fig:7}
\end{figure*}
Validation evaluation reorders several of these results. Capacity buys the base model nothing measurable, since YOLO26 spans 0.773 to 0.778 mAP50 across a 25-fold parameter range on the selection split and orders non-monotonically under validation evaluation, its largest arm reaching only 0.719, whereas YOLO26-RD-x attains the highest validation mAP50 in the study at 0.761. At the small scale the proposed model reaches parity with the best base model at 61\% of its parameters, 0.737 validation mAP50 against 0.735, a difference inside the resolution limit. The external family provides the least favourable comparison on the validation split, where YOLOv12-x returns the highest mAP50-95 recorded here, 0.494 against 0.484 for YOLO26-RD-x, an ordering opposite to the selection-split one in which the same two arms return 0.481 and 0.495. The unanimous lead over both reference families at scales m, l and x is therefore a selection-split result, and it is at scale x that the validation split declines to reproduce it. The claim the two splits jointly support is the one against the recipe-matched YOLO26 baseline; the claim against the external family is supported on the split that selected the checkpoints and, at the top of the ladder, contradicted on the split that selected nothing.

Figure 8 replots the three families against measured inference latency on an RTX 5090, and the ordering follows that of Figure 7 with the abscissa expressed in the units a deployment actually budgets. Compilation to a TensorRT FP16 engine reduces inference by a factor of 8.3 to 11.3 and places all fifteen models between 1.22 and 3.31 ms per image, so the entire study occupies a range of about two milliseconds. Below 1.5 ms the unmodified YOLO26 holds the accuracy frontier, returning 0.773 mAP50 at 1.22 ms and 0.777 at 1.27 ms. From 1.64 ms upward the proposed architecture holds it instead, at 0.787 mAP50 for scale s, 0.794 for m and 0.809 for l, and it retains the mAP50 frontier at every latency beyond that point. On the stricter metric it takes the frontier only from 1.89 ms upward, since scale s returns 0.469 mAP50-95, which the cheaper YOLO26-m already exceeds at 0.474 and 1.27 ms. Two comparisons within it are strict dominations rather than trades: YOLO26-RD-m returns 0.794 mAP50 and 0.488 mAP50-95 at 1.89 ms against 0.782 and 0.482 for YOLOv12-l at 2.55 ms, and YOLO26-RD-l attains 0.809 mAP50 at 2.64 ms, a value no arm of the external family reaches at any latency. The highest accuracy in the study belongs to that same arm on both metrics, at 0.809 mAP50 and 0.497 mAP50-95, while the largest external arm, YOLOv12-x, returns 0.792 and 0.481 at 3.31 ms and 59.05M parameters and therefore neither exceeds YOLO26-RD-x at equal latency nor recovers the deficit of the middle range. The scale-x arms nonetheless lie outside the training budget the rest of the study used, since those of all three families require two RTX 5090s to hold batch 32 rather than one. Two stages excluded from the abscissa qualify all of the above. Preprocessing costs 2.1 to 2.5 ms for every model in the study and therefore exceeds the inference stage itself for every arm up to scale m, and postprocessing separates the families independently of the abscissa, at 0.35 to 0.49 ms for the NMS-free YOLO26 and YOLO26-RD heads against 0.56 to 1.26 ms for YOLOv12.

\begin{figure*}[!t]
\centering
\includegraphics[width=0.82\textwidth]{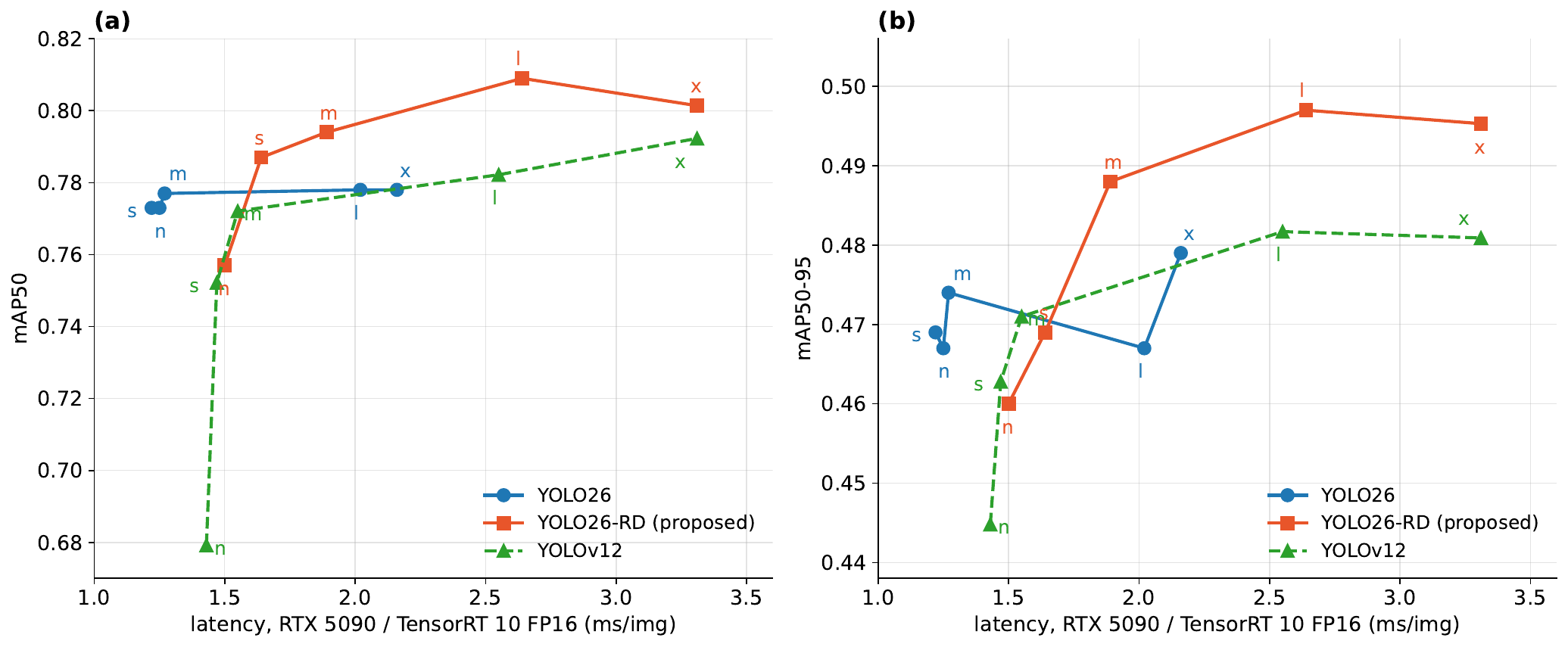}
\caption{Accuracy against TensorRT FP16 inference latency on an RTX 5090, at scales n/s/m/l/x: YOLO26 (blue circles), YOLO26-RD (orange squares), YOLOv12 (green triangles, dashed). \emph{(a)} mAP50. \emph{(b)} mAP50-95. All points are selection-split values, as in Figure 7; protocol and excluded stages are given in the text above.}
\label{fig:8}
\end{figure*}
Figure 9 repeats the measurement on an RTX 5070 and an RTX 5060 as TensorRT FP16 engines at batch one and 640² input. The three series are not quantitatively comparable: the RTX 5090 and RTX 5060 time the inference stage alone whereas the RTX 5070 series is end-to-end, and they come from different harnesses, which shows in the figure itself, since on the RTX 5070 the inference stage alone takes 0.84 ms at scale n against 1.50 ms on the more capable RTX 5090. Within each series the rank order of the five scales is identical but the spacing is not, the ratio of the scale-x arm to the scale-n arm being 2.2, 4.4 and 7.3 respectively, so a budget expressed in milliseconds selects a different scale on each device. Latency is deterministic per frame, the ninetieth percentile exceeding the median by at most 0.01 ms at every scale measured, which is the property the NMS-free head was adopted to provide.

\begin{figure*}[!t]
\centering
\includegraphics[width=0.95\textwidth]{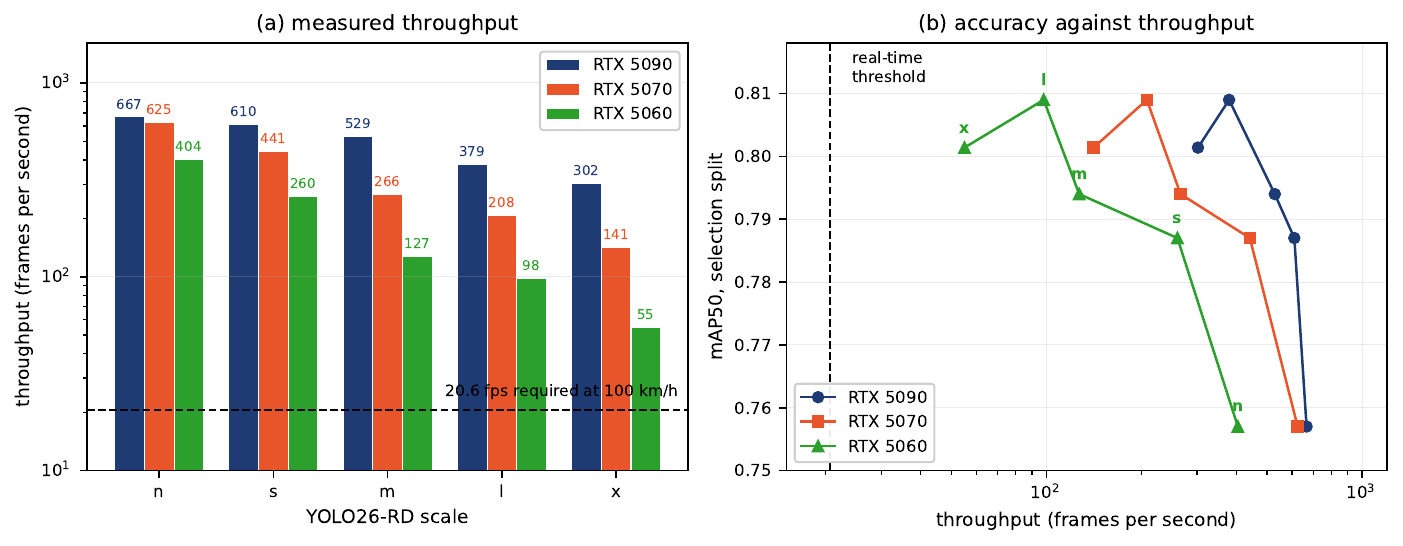}
\caption{Throughput of YOLO26-RD across deployment hardware, as TensorRT FP16 engines at batch one and 640² input. \emph{(a)} Measured throughput by scale on three devices, each bar labelled with its value; the dashed line marks the 20.6 frames per second that a 100 km/h survey requires at half-frame overlap (Section 8). \emph{(b)} The same throughput against selection-split mAP50, with the five scales labelled on the RTX 5060 series; all three series carry the same checkpoints, so the curves differ from one another only by a horizontal shift. Latency in milliseconds is the reciprocal of the plotted throughput. The RTX 5090 and RTX 5060 series time the inference stage alone and the RTX 5070 series is end-to-end, and the three were measured under different benchmarking harnesses, so they are not quantitatively comparable to one another; see the text above. Parameter counts and the accompanying mAP50-95 figures are given in Table 6.}
\label{fig:9}
\end{figure*}
\subsection{Qualitative comparison of activation maps}\label{qualitative-comparison-of-activation-maps}

Figure 10 traces channel-mean activation maps through matched stages of the two trained scale-l checkpoints on three selection-split frames, one per class. The six columns run from the network input at 640², which for YOLO26-RD-l is the LearnableContrast output, through the stem P2/4 and the backbone P3/8 and P4/16 stages, to the head-P3 and head-P4 tensors that feed the detection layer. Each map is normalized independently and rendered on a jet scale, so dark blue marks the weakest response within a map and yellow and red the strongest; colour shows where a stage responds, not how strongly. The input column establishes that LearnableContrast is non-identity at inference. At the stem the annotated geometry appears as shape in both models, the patch perimeter of panel (a) as a closed cyan contour and the two cracks of panel (b) as narrow vertical lines. The head-P3 column, the finest scale at which the proposed model predicts once the stride-4 head is removed, carries the comparison. In panel (b) the baseline map is a broad cyan to yellow field filling the unannotated centre, averaging 0.38 outside the annotation against 0.30 within, so its warmest colour lies on plain asphalt rather than on the two labelled cracks; the proposed model holds that centre at dark blue, 0.25 against 0.40, and the surviving warm colour takes the shape of the distress as two narrow vertical stripes at the annotated positions. The pattern is that the proposed architecture separates distress from pavement at the stride-8 level by darkening the background rather than by brightening the peaks. Three hand-inspected frames cannot establish this as a property of the architecture, and one contrary case appears in the head-P4 column of panel (a).

\begin{figure*}[!t]
\centering
\includegraphics[width=0.95\textwidth]{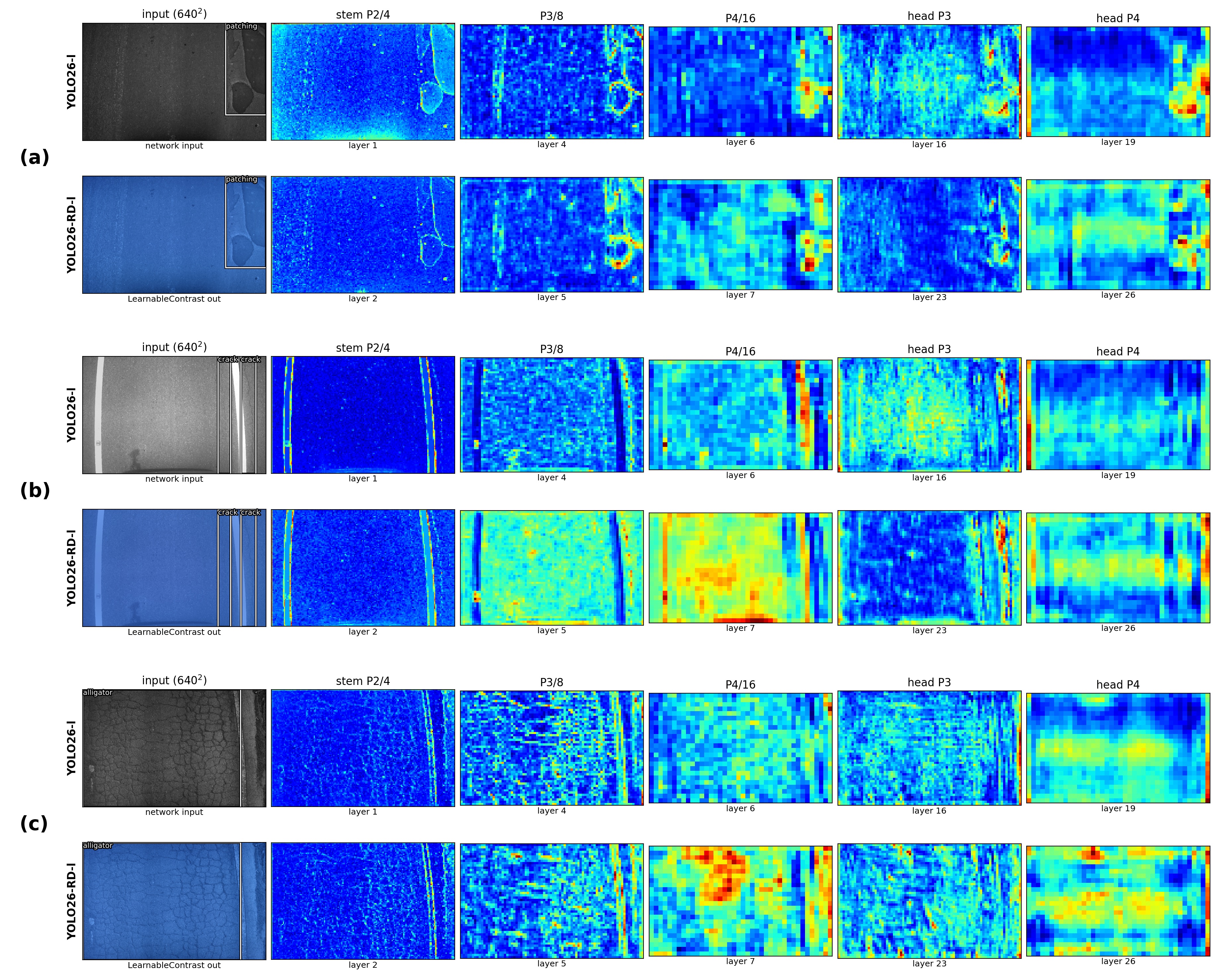}
\caption{Channel-mean activation maps for YOLO26-l (upper row of each panel) and YOLO26-RD-l (lower row) on three selection-split frames: \emph{(a)} a patching region, \emph{(b)} two full-height longitudinal cracks, \emph{(c)} an alligator-cracked area. Columns, stage indices and colour scale are specified in the text above.}
\label{fig:10}
\end{figure*}
\subsection{Representative failure cases}\label{representative-failure-cases}

Figure 11 examines four failures of the YOLO26-RD scale-l checkpoint on the selection split, retained from a label-guided screen of twelve candidates. This is a qualitative probe rather than a second evaluation, so it cannot estimate how often any mechanism occurs. Ground-truth boxes are classified by the matching logic of Section 3.2. Panel (a) isolates a boundary effect: the model detects a large alligator region at IoU 0.82 but misses a smaller one truncated by the lower image boundary, two instances sharing class and appearance. Panel (b) is a single narrow crack occupying 0.3\% of image area, lying beside a stronger construction joint and drawing no prediction. Panel (c) contains three patching regions and no predictions, two of them long, narrow repairs clipped by the right boundary whose appearance is dominated by seams rather than by a compact patch interior. Panel (d) shows the failure produced by crowding: among seven crossing crack annotations, two are hits, two receive only partial responses and three are missed, the predictions merging portions of intersecting cracks into boxes that do not follow the annotation partition. Residual errors therefore concentrate where the visible region is clipped, faint, atypically shaped or split ambiguously among intersecting boxes, and increasing model scale alone is unlikely to remove them.

\begin{figure*}[!t]
\centering
\includegraphics[width=0.95\textwidth]{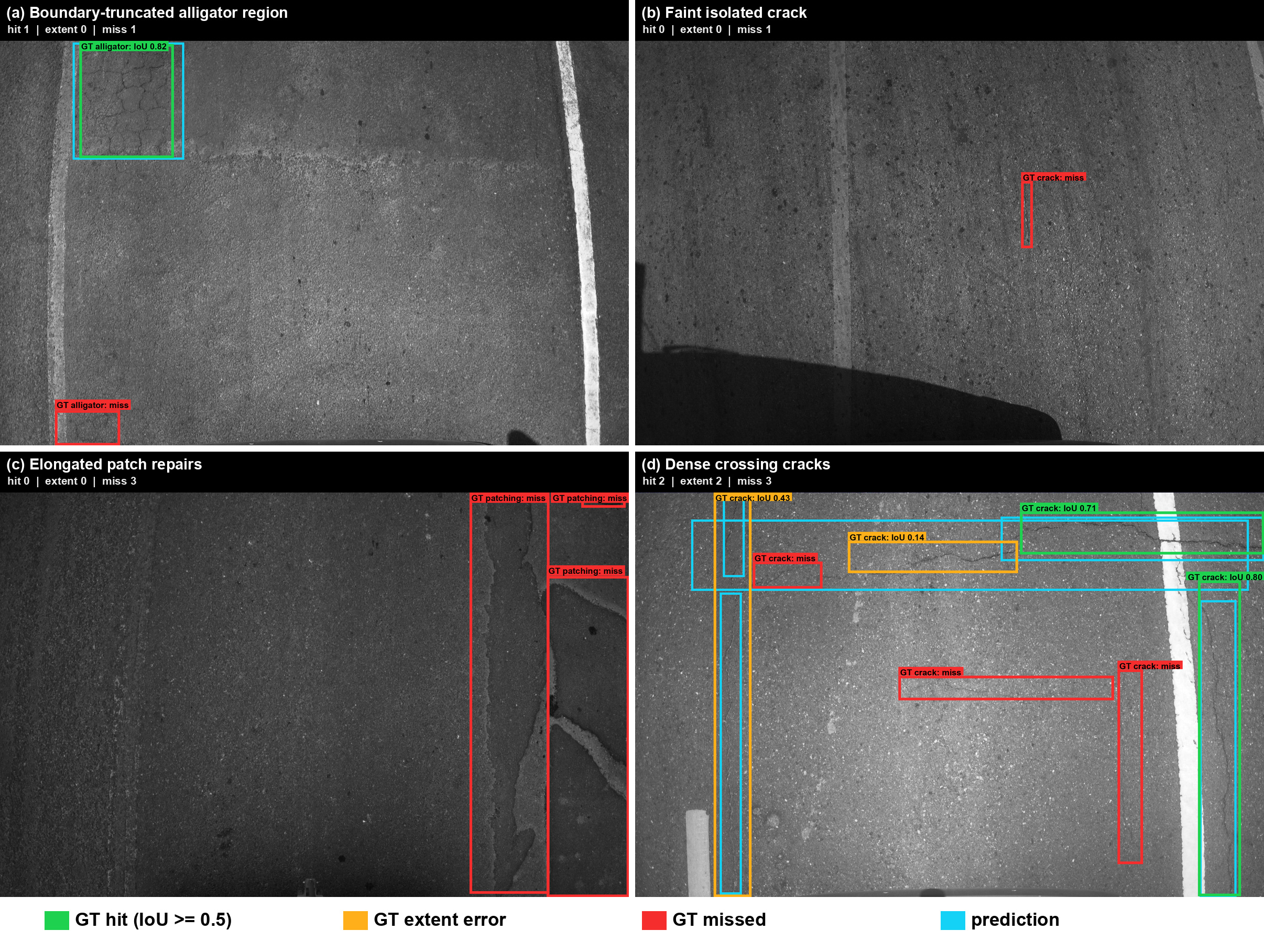}
\caption{Four representative failures of YOLO26-RD-l on the selection split: \emph{(a)} a small boundary-truncated alligator region missed beside a correctly detected larger one; \emph{(b)} a faint isolated crack with no prediction; \emph{(c)} three elongated or boundary-clipped patching regions with no prediction; \emph{(d)} a dense seven-crack scene. Ground-truth boxes are coloured by outcome, green for a hit, amber for an extent error and red for a miss, and predictions are cyan. Selection procedure and matching thresholds are given in the text above.}
\label{fig:11}
\end{figure*}
\section{Discussion}\label{discussion}

The design decision with the strongest empirical support in this study is the removal of a detection level, and it was determined by measurement rather than by architectural reasoning. The audit of Section 3.2 established that 1.28\% of instances are COCO-small at the 640² training resolution while 70.37\% are large, since a linear crack is annotated as the region enclosing it, and that a stride-4 detection level would supply 75.3\% of all anchors. The failure decomposition agreed: complete misses dominate every class, at 10\%, 16\% and 34\% of alligator, crack and patching ground truth, whereas localization error accounts for approximately 7\% of crack instances. Removing that detection level while retaining its features in the neck therefore redirects assignment budget toward the error mode that dominates, and it is the only modification reported here whose sign is preserved on both splits. Its identification incurred no additional training, whereas each ablation arm required approximately 4.7 GPU-hours at scale l.

That result generalises beyond this dataset only to the extent that the labelling convention does. The instances here are large because a linear crack is annotated as an elongated region; on a corpus annotated at pixel level, or on the crack-level benchmarks that motivated the detectors of Table 1, the same measurement could support the conventional recipe. What transfers is the procedure rather than the conclusion, since none of the nine pavement detectors surveyed in Table 1 reports measuring its own box geometry before adding small-object machinery.

Each proposed module improves the base architecture on its own, and together they give the best model of the six: every arm in Table 5 exceeds the YOLO26-l baseline on both aggregate metrics, and the full configuration reaches 0.809 mAP50 and 0.497 mAP50-95 while leading all three per-class entries. What the grid does not support is the attribution of that gain to either module separately. Relative to the baseline the contrast stem contributes 0.004 mAP50 and gated downsampling at two locations 0.016, whereas jointly they contribute 0.031, and the stem contributes −0.004 at one gated location and +0.015 at two, so the sign of its measured effect is conditional on the other module. A single run per cell cannot separate that interaction from noise, four of the six adjacent-cell contrasts fall below the ±0.015 resolution limit, and no arm of the grid was evaluated on the validation split. The modules' premises may also be under-expressed in this corpus, since uniformly lit down-facing survey imagery is the least demanding case on which to measure a contrast correction.

The proposed architecture outperforms both reference families, and it does so as a family rather than at one tuned point. Averaged over the five scales it returns 0.790 mAP50 and 0.482 mAP50-95 against 0.776 and 0.471 for a recipe-matched YOLO26 and 0.755 and 0.468 for a recipe-matched YOLOv12; it gives the higher mAP50 than both at every scale from s upward; at scales m, l and x it leads both on mAP50 and mAP50-95 alike, in every one of the twelve comparisons available; and the most accurate model of the fifteen trained is YOLO26-RD-l on both metrics, at 0.809 mAP50 and 0.497 mAP50-95. Where that advantage ends, its boundary is set by capacity, and it lies at the bottom of the ladder rather than in the middle of it. Figures 7 and 8 show three families differing in slope: the base model is almost insensitive to scale, spanning 0.773 to 0.778 mAP50 across a 25-fold parameter range, the external family is strongly parameter-sensitive over the same range at 0.679 to 0.792, and the proposed model lies between them. The proposed model's curve consequently crosses each reference exactly once, between scales n and s, and it is only there that the choice of architecture turns on the budget: at scale n the base model is the most accurate of the three, whereas from scale s upward the proposed model leads both references on mAP50 at every scale and from scale m upward on both aggregate metrics at every scale, a sweep of twelve pairwise comparisons decided in one direction. The latency abscissa expresses the same ordering in deployable units, YOLO26-RD-m exceeding YOLOv12-l on both metrics at 1.89 ms against 2.55 ms, and YOLO26-RD-l reaching 0.809 mAP50 at 2.64 ms, which no arm of the external family attains at any latency. Two qualifications bound how far that sweep can be pressed. Three of the twelve margins, mAP50-95 against YOLO26 at scale m and both metrics against YOLOv12 at scale x, fall inside the ±0.015 resolution limit and are individually inconclusive, so the case for them rests on the consistency of the set rather than on their own size. And the sweep is a selection-split result: the validation split reproduces the ordering against the base model but reverses it against the external family at scale x, which is the reason the comparison against YOLOv12 is reported here as the weaker of the two claims.

The scale-l comparison is the one the released model rests on. Every arm ran the same schedule, so the pair is a controlled architectural comparison. On the test split the proposed model leads on both aggregate metrics and in all six per-class entries, the operating point moving toward sensitivity as recall rises and precision falls, which is the direction the audit called for. At scale s the same change raises recall further, so the mechanism is capacity-dependent in a way this study measures but does not explain.

The ceiling on the crack class is attributable to the data rather than to the network, and no architecture evaluated here alters it. The ratio of AP50-95 to AP50 is 0.44 for crack against 0.80 for patching, characteristic of instances detected but loosely bounded. The cause is geometric: for a median crack box of aspect 10.3:1, overlap is governed almost entirely by the short axis, so a 10 px error in width reduces IoU to 0.771 whereas the same error along the length is negligible. The crack line itself is 0.5 to 1.1 px wide at the training resolution and the surrounding box edge is annotator-chosen, so the regression target is ambiguous in the dimension that determines the score, and oriented boxes offer no remedy since 94.5\% of crack boxes are already axis-elongated. Further architecture search optimises the wrong variable; a disclosed relabelling pass does not.

Three properties of the compiled system matter more for deployment than the margin in mAP, and all derive from the base template rather than from the modules proposed here. The head is NMS-free, so prediction count and postprocessing time are deterministic, at 0.35 to 0.49 ms against 0.56 to 1.26 ms for the NMS-based external heads. Compiled rather than eager latency is the relevant quantity, since a TensorRT FP16 engine reduces inference by a factor of 8.3 to 11.3, after which preprocessing at 2.1 to 2.5 ms exceeds the inference stage for every arm up to scale m. For the on-board deployment of Section 1 none of these costs is binding, since the released scale-l model sustains 98 frames per second on an entry-level RTX 5060 against the 21 required at 100 km/h, and returning per-frame records rather than imagery reduces the transmitted volume from 661 KB to approximately 150 bytes.

Several conditions bound these results. The dataset cannot separate differences below approximately ±0.015 mAP50 at 724 selection and 686 validation instances. The module grid was scored on one split and omits its module-free cell, every arm was trained at 640² under a single seed, and from-scratch training may understate all arms relative to pretrained practice. Measured precision is a lower bound for every arm, since most of the 113 confident false positives examined proved to be unlabelled distress. The stride-4 removal is specific to region-level annotation and should not be assumed to transfer to pixel-thin crack segmentation. The deployment figures rest on a ground sampling distance inferred from lane geometry rather than from a calibrated camera, on three devices timed under harnesses reporting different stages, and on no measurement of power draw, sustained thermal behaviour or vibration tolerance.

\section{Conclusion}\label{conclusion}

This paper examined whether the small-object specialisation conventional in pavement-distress detectors is warranted by the data those detectors are trained on. An audit of a 7,618-image road-survey corpus annotated at region level shows that it is not: 1.28\% of instances are small at the 640² training resolution and 70.37\% are large, a stride-4 detection level would supply 75.3\% of all anchors, and complete misses rather than localization errors dominate the failure decomposition of a trained baseline. YOLO26-RD therefore retains the stride-4 branch as neck features but carries no detection level there, and adds a per-tile learned contrast stem and an edge-gated lossless downsampler. At scale l it reaches 0.809 mAP50 and 0.497 mAP50-95 against 0.778 and 0.467 for the YOLO26 reference, leading in every per-class entry, with recall rising and precision falling as the audit predicted, at a cost of 40.5\% more parameters and 30.7\% more compiled inference time.

YOLO26-RD is the most accurate of the three architectures compared in this study, and YOLO26-RD-l is the most accurate single detector among the fifteen models trained. Fifteen arms, five scales of each family, were trained from scratch under one recipe on identical data, splits and resolution, so what separates them is architecture; averaged over its five scales the proposed family returns 0.790 mAP50 and 0.482 mAP50-95 against 0.776 and 0.471 for YOLO26 and 0.755 and 0.468 for YOLOv12. The ordering holds scale by scale and not only in the mean, since the proposed model returns the higher mAP50 than both references at every scale from s upward and leads them on mAP50 and mAP50-95 alike at scales m, l and x, twelve pairwise comparisons decided without a single exception. Its scale-l arm gives 0.809 mAP50 and 0.497 mAP50-95, the highest values recorded on either metric anywhere in the study, and no arm of either reference family reaches either figure at any scale. That arm is also the more efficient model in the comparison rather than merely the larger one, since it exceeds both scale-x references, YOLO26-x at 0.778 and 0.479 and YOLOv12-x at 0.792 and 0.481, while carrying about 61\% of their parameters and with all four margins outside the ±0.015 resolution limit, and on the latency abscissa it reaches 0.809 mAP50 at 2.64 ms per frame, a value no arm of the external family attains at any latency. Two conditions qualify the claim without displacing it: three of the twelve scale-wise margins fall inside that resolution limit and rest on the consistency of the set rather than on their own size, and the held-out split reproduces the ordering against YOLO26 but not against YOLOv12 at the largest scale. What the evidence supports is therefore a family-level result rather than one tuned configuration, with the released scale-l model as the best detector this study measured.

Three findings bound those results. First, the module grid does not decompose into independent contributions: the two modules add 0.004 and 0.016 mAP50 separately but 0.031 jointly, and the sign of the contrast stem's effect is conditional on the other module, so neither has a single effect size on these data. Second, the advantage depends on capacity and is absent only at the smallest scale trained. The base model is almost insensitive to scale, spanning 0.773 to 0.778 mAP50 across a 25-fold parameter range, whereas the external family is strongly parameter-sensitive over the same range, spanning 0.679 to 0.792; the proposed model's curve crosses each reference once, between scales n and s, so the base model is preferable at scale n, while from scale s upward the proposed model leads both references on mAP50 and from scale m upward leads them on both aggregate metrics at every scale, winning all twelve pairwise comparisons at m, l and x. Three of those twelve margins fall inside the resolution limit, and the validation split, which reproduces the ordering against the base model, does not reproduce the lead over the external family at the largest scale. Third, the crack class is bounded by annotation geometry rather than by capacity, since overlap on a 10:1 box is governed by its short axis, where a 10 px error reduces IoU to 0.771, and most confident false positives examined proved to be unlabelled distress, which makes measured precision a lower bound.

For on-board deployment the decisive properties derive from the NMS-free base template rather than from the proposed modules. Prediction count and per-frame latency are deterministic, and the released model sustains 98 frames per second on an entry-level accelerator against the 21 a 100 km/h survey requires, so accuracy rather than latency selects the scale; transmitting per-frame detection records rather than imagery, about 150 bytes against 661 KB, removes the link as a constraint on network-scale operation.

The transferable contribution of this work is methodological. Auditing annotation geometry and failure modes before designing for them reversed the architecture this study set out to build, and it is the step the pavement-distress literature omits. Where an architectural gain is claimed, the split that selected the checkpoint cannot establish it, and success criteria fixed before each run are what obliged this study to report as failures the arms such a split would have reported as successes.

\subsection{Future work}\label{future-work}

The results reported here define four directions for subsequent research. The first concerns the two proposed modules, whose effects Section 7.1 shows to be conditional on one another, the sign of the contrast stem's contribution changing with the other module's setting. Replication across initialisations, together with the module-free cell of the grid, would determine whether that interaction is structural; scoring the existing intermediate checkpoints on a second split extends the question to generalisation at no additional training cost.

The second concerns capacity: the same modification yields a sensitivity gain at scale s and a localisation gain at scale l, and what governs that dependence is not identified here. The third concerns the data. A disclosed relabelling pass over the confident false positives of Section 3.2 would raise the crack-class ceiling and convert measured precision into an estimate, repeating the audit on corpora annotated at crack or pixel level would establish how far the detection-scale finding carries, and an illumination-stressed benchmark would supply the regime in which a local contrast correction should matter most. The fourth concerns deployment: instrumenting the complete pipeline of Figure 1 in a working vehicle on embedded hardware would characterise the survey system rather than the detector, the quantities of interest being sustained frame rate, throughput under thermal load, power envelope, storage and link behaviour in the field.

\section{Reproducibility}\label{reproducibility}

Code and trained weights are available at https://github.com/Sompote/YOLO26RD. Fork: Ultralytics 8.4.115 with \texttt{EdgeSPD} and \texttt{LearnableContrast} in \texttt{ultralytics/nn/modules/conv.py}, registered in \texttt{nn/tasks.py}. Training: single RTX 5090 for every arm except the scale-x arms of all three families, which require 2 × RTX 5090 to hold batch 32, \texttt{imgsz=640,\ batch=32,\ epochs=150,\ optimizer=auto\ (MuSGD\ lr\ 0.01),\ cos\_lr,\ mosaic=0.5,\ close\_mosaic=30,\ flipud=0.5,\ fliplr=0.5,\ seed=0}, from scratch. Evaluation: Ultralytics \texttt{DetMetrics}, fitness-selected checkpoint, \texttt{imgsz=640}, once per model per split. Reference families: the YOLO26 and YOLOv12 arms of Table 6 were trained and evaluated in this same fork, the YOLOv12 arms instantiated from its \texttt{ultralytics/cfg/models/12/} configurations with no modification to the YOLOv12 source. Audit scripts (geometry statistics by resolution, split-leakage check, failure decomposition, IoU-sensitivity analysis, contact-sheet generation) are released with the code, together with the module-level prediction list and the decision rules that were fixed before each run.

\section{Data Availability Statement}\label{data-availability-statement}

The modified Ultralytics fork containing both modules, the training configurations, the released checkpoints and the audit scripts (geometry statistics by resolution, split-leakage check, failure decomposition, IoU-sensitivity analysis, contact-sheet generation) are available at https://github.com/Sompote/YOLO26RD. The survey imagery was collected under a commercial pavement-inspection contract and cannot be redistributed in full; the per-instance annotation-geometry statistics on which the audit rests are included with the released code.

\section{Author Contributions}\label{author-contributions}

SY: conceptualization, methodology, software, formal analysis, investigation, writing --- original draft, writing --- review and editing. PC: data curation, resources, validation, writing --- review and editing. HS: data curation, annotation and annotation review, validation. TY: resources, data acquisition, project administration. All authors read and approved the submitted version.

\section{Funding}\label{funding}

This research received no external funding.

\section{Conflict of Interest}\label{conflict-of-interest}

Authors Pawarotorn Chaipetch, Hathairat Samaikul and Theerayut Yonseng are employed by Infraplus Co., Ltd., which supplied the imagery and annotations used in this study. The remaining author declares that the research was conducted in the absence of any commercial or financial relationships that could be construed as a potential conflict of interest.

\section{Appendix A. Supplementary tables}\label{appendix-a.-supplementary-tables}

\begin{table*}[!t]
\centering
\small
\setlength{\tabcolsep}{6pt}
\caption{Data audit of the imagery and of the unmodified YOLO26-l reference. Size classes follow the COCO convention applied after letterboxing. The failure decomposition is exhaustive per ground-truth instance under one-to-one matching over the 724 test-split instances, at an operating point returning precision 0.828, recall 0.675, mAP50 0.704 and mAP50-95 0.421, which is not the framework's default and is therefore not interchangeable with the dataset-level metrics reported elsewhere.}
\label{tab:A1}
\begin{tabularx}{\textwidth}{XXX}
\toprule
audit step & measurement & implication \\
\midrule
Instance size (640² input) & 1.28\% small, 28.34\% medium, 70.37\% large; smallest instance √area 14.4 px; small share 0.02\% at 1024² and 0.01\% at 1280² & the small-object premise fails at every resolution considered \\
Annotated box geometry & median crack box 138 × 1,476 px at aspect 9.5:1 (p95 17.2:1), spanning about 90\% of frame height; alligator median √area 1,020 px; patching median aspect 0.98 & instances are thin, not small, and no class occupies a small-object regime \\
Anchor allocation (640² input) & P3, P4 and P5 supply 6,400, 1,600 and 400 anchors, 8,400 in total; a stride-4 level adds 25,600, or 75.3\% of all anchors & a stride-4 detection level is not warranted on this data \\
Failure decomposition (YOLO26-l, 724 instances) & missed entirely 25 of 241 alligator (10\%), 67 of 418 crack (16\%), 22 of 65 patching (34\%); localization error 30 of 418 crack (7\%) and 0 of 65 patching; misclassified 2 in the whole split & the deficit is sensitivity, not box regression \\
Geometry shift, train to test (crack) & median crack spans 0.90 of frame height in training against 0.51 in the test split; transverse share 18.9\% against 38.3\% & part of the crack miss rate is a geometry difference between the splits, not a capacity limit \\
Label completeness (confidence ≥ 0.5, IoU < 0.1) & the majority of confident unmatched detections are real but unlabelled distress; 74 of the 113 false positives at IoU < 0.3 belong to \texttt{crack} & measured precision is a lower bound, depressed equally for every arm \\
\bottomrule
\end{tabularx}
\end{table*}
\begin{table*}[!t]
\centering
\footnotesize
\setlength{\tabcolsep}{4pt}
\caption{YOLO26-RD layer-by-layer specification at scale l (depth 1.00, width 1.00, maximum 512 channels). Parameter counts are per layer for the unfused model at \texttt{nc = 3}. Groups of three layers are given as one row with their summed count, the \texttt{n} and \texttt{c\_out} entries being those of the fusion block terminating the group.}
\label{tab:A2}
\begin{tabularx}{\textwidth}{lcXccccX}
\toprule
\# & from & module & n & c\_out & params & stride & role \\
\midrule
0 & −1 & \textbf{LearnableContrast} & 1 & 3 & 494 & /1 & tile-wise contrast stem (proposed) \\
1 & −1 & Conv 3×3 s2 & 1 & 64 & 1,856 & /2 & P1 \\
2 & −1 & Conv 3×3 s2 & 1 & 128 & 73,984 & /4 & P2 \\
3 & −1 & C3k2 & 2 & 256 & 173,824 & /4 &  \\
4 & −1 & \textbf{EdgeSPD} & 1 & 256 & 2,359,810 & /8 & P3 edge-gated lossless downsample (proposed) \\
5 & −1 & C3k2 & 2 & 512 & 691,712 & /8 & backbone P3 out \\
6 & −1 & \textbf{EdgeSPD} & 1 & 512 & 9,438,210 & /16 & P4 edge-gated lossless downsample (proposed) \\
7 & −1 & A2C2f & 2 & 512 & 2,268,568 & /16 & backbone P4 out (area attention) \\
8 & −1 & Conv 3×3 s2 & 1 & 512 & 2,360,320 & /32 & P5 \\
9 & −1 & C3k2 & 2 & 512 & 2,234,368 & /32 &  \\
10 & −1 & SPPF & 1 & 512 & 656,896 & /32 &  \\
11 & −1 & C2PSA & 2 & 512 & 1,455,616 & /32 & backbone P5 out \\
12–14 & 11, 7 & Upsample · Concat · C3k2 & 2 & 512 & 2,496,512 & /16 & top-down T4 \\
15–17 & 14, 5 & Upsample · Concat · A2C2f & 2 & 256 & 592,640 & /8 & top-down T3 (area attention) \\
18–20 & 17, 3 & Upsample · Concat · C3k2 & 2 & 128 & 189,952 & /4 & top-down T2, \textbf{P2 features, fused not detected} \\
21–23 & 20, 17, \textbf{5} & Conv s2 · Concat · C3k2 & 2 & 256 & 871,680 & /8 & bottom-up B3 (+ backbone-P3 skip) \\
24–26 & 23, 14, \textbf{7} & Conv s2 · Concat · C3k2 & 2 & 512 & 3,217,920 & /16 & bottom-up B4 (+ backbone-P4 skip) \\
27–29 & 26, 11 & Conv s2 · Concat · C3k2 & 1 & 512 & 4,335,104 & /32 & bottom-up B5 \\
30 & \textbf{23, 26, 29} & Detect (end2end, \texttt{reg\_max} 1) & 1 & — & 2,803,242 & — & \textbf{Detect(P3, P4, P5)}, input channels (256, 512, 512) \\
\bottomrule
\end{tabularx}
\end{table*}
\section{Appendix B. Module definitions}\label{appendix-b.-module-definitions}

\textbf{LearnableContrast.} For an input \(x\) and a tile grid \(g\), the module computes:

Given \(x \in [0,1]^{B \times C \times H \times W}\) and a tile grid \(g = \min(8, H, W)\):

\begin{itemize}
\tightlist
\item
  Tile statistics: \(\mu = \mathrm{AvgPool}_g(x)\), \(\sigma = \sqrt{\mathrm{AvgPool}_g(x^2) - \mu^2 + \varepsilon}\)
\item
  Correction prediction: \((\Delta\gamma, \Delta\alpha) = f([\mu;\, \sigma])\), where \(f\) is \(\mathrm{Conv}_{3\times3}(2C{\to}r) \to \mathrm{SiLU} \to \mathrm{Conv}_{1\times1}(r{\to}2C)\), with the final layer zero-initialized so the module is exactly the identity at initialization
\item
  Bounded corrections: \(\gamma = 1 + \tanh(\Delta\gamma) \in (0,2)\), \(\quad\alpha = 1 + \tanh(\Delta\alpha) \in (0,2)\)
\item
  Bilinear upsampling of \(\gamma, \alpha\) to \(H \times W\) (the analogue of CLAHE's cross-tile blending), then
\end{itemize}

\[y = \mathrm{clip}\left(x^{\gamma} \cdot \alpha,\; 0,\; 1\right)\]

\textbf{EdgeSPD.} For an input feature map \(x\), the module computes:

Given \(x \in \mathbb{R}^{B \times C \times H \times W}\):

\begin{itemize}
\tightlist
\item
  Gradient prior: \(e = \sqrt{(S_x * \bar{x})^2 + (S_y * \bar{x})^2}\), with \(S_x, S_y\) fixed Sobel kernels applied to the channel-mean map \(\bar{x}\)
\item
  Gate: \(x' = x \cdot \left(1 + \sigma(w e + b)\right)\), a single learnable 1×1 convolution on the gradient magnitude (2 parameters)
\item
  Space-to-depth: the four pixel-decimated sub-maps of \(x'\) are concatenated channel-wise (\(C \to 4C\), \(H, W \to H/2, W/2\)) and fused by a 3×3 convolution, exactly SPD-Conv applied to the gated tensor.
\end{itemize}

\end{document}